\documentclass{article}

\usepackage{PRIMEarxiv}

\usepackage[utf8]{inputenc} 
\usepackage[T1]{fontenc}    
\usepackage{hyperref}       
\usepackage{url}            
\usepackage{booktabs}       
\usepackage{amsfonts}       
\usepackage{nicefrac}       
\usepackage{microtype}      
\usepackage{lipsum}
\usepackage{xcolor} 
\usepackage{xcolor} 
\usepackage{fancyhdr}       
\usepackage{graphicx}       
\graphicspath{{media/}}     

\usepackage{subcaption}

\newcommand{\lead}{{Team Lead}}
\newcommand{\cog}{\textit{cog}}
\newcommand{\op}{{Operator-cog}}
\newcommand{\ana}{{Analyst-cog}}
\newcommand{\gp}{{GP-cog}}
\newcommand{\writer}{{Writer-cog}}
\newcommand{\audit}{{Audit-cog}}
\newcommand{\percent}{$\%$ } 
\newcommand{\expone}{\textit{Exp1}}
\newcommand{\exptwo}{\textit{Exp2}}
\newcommand{\expthree}{\textit{Exp3}}

\title{Auditable Human–Agent Teams for Autonomous X-ray Experimentation
}

\author{
  Gunik Luthra\\
  Center for Functional Nanomaterials\\
  Brookhaven National Laboratory \\
  Upton, NY 11973, USA\\
  \And
  Ronald J. Pandolfi \\ 
  Applied Mathematics \\ and Computational Research\\
  Lawrence Berkeley National Laboratory \\
  Berkeley, CA 94720, USA \\
  \And 
  Benjamin T. Diroll \\ 
  Center for Nanoscale Materials\\
  Argonne National Laboratory \\
  Lemont, IL 60439, USA 
  \And
  Ricky Ronquillo \\ 
  University of Pennsylvania \\
  Philadelphia, PA 19104, USA \\ 
  \And
  Ruipeng Li \\ 
  National Synchrotron Light Source II\\
  Brookhaven National Laboratory \\
  Upton, NY 11973, USA\\
  \And
  Christopher B. Murray \\ 
  University of Pennsylvania \\
  Philadelphia, PA 19104, USA \\ 
  \And
  Marcus M. Noack \\ 
  Applied Mathematics \\ and Computational Research\\
  Lawrence Berkeley National Laboratory \\
  Berkeley, CA 94720, USA \\
  \And
  Kevin G. Yager \\
  Center for Functional Nanomaterials\\
  Brookhaven National Laboratory \\
  Upton, NY 11973, USA\\
  \And
  Esther H. R. Tsai \\
  Center for Functional Nanomaterials\\
  Brookhaven National Laboratory \\
  Upton, NY 11973, USA\\
  \texttt{etsai@bnl.gov} \\
}

\begin{document}
\maketitle

\begin{abstract}
Closed-loop autonomous experimentation can accelerate scientific discovery, but implementation effort limits accessibility. Agentic systems may lower this barrier, although sustained operation on physical instruments remains largely unexplored. Here, using coding-agent infrastructure grounded in validated scientific methods, we demonstrate autonomous synchrotron X-ray experiments on nanoparticle superlattices under evolving, human-defined objectives. 
Complete human--agent and agent--agent transcripts enable an audit that reconstructs decision provenance and characterizes the division of cognitive labor.
Across three experiments, 61--67\% of messages involved interpretation or judgment, while the share of methodological and scientific contributions increased from 46\% in the setup-focused first experiment to 56--61\% in later experiments. 
Provenance showed that {59--68\%} of error corrections involved multiple participants and that about half of all decisions incorporated the input of another participant. These results may guide design principles for multi-agent scientific systems and inform future agent harnesses, model training, and accessible autonomous experimentation.

\end{abstract}


\section{Introduction}

\textbf{AI for Science}

AI agents are increasingly capable of executing substantial portions of the scientific workflow.
The AI Scientist~\cite{AIScientist2024,AIScientistV2_2025,Lu2026NatureAutomation} has demonstrated increasingly general automation of computational research, spanning idea generation, coding, experiments, analysis, and manuscript writing. Systems based on large language models (LLMs) can also help scientists search and synthesize findings across large bodies of prior work~\cite{asai2026synthesizing}.
AI agents are also being evaluated for scientific peer review, where they can identify issues missed by individual reviewers while still benefiting from human oversight~\cite{kim2026limits}.

Domain-specific systems have coordinated atomistic simulations and evaluated experimental designs through first-principles calculations~\cite{vriza2025multi_agentic,hill2026grace}, while coding agents have reconstructed scientific software from publications, reproduced computational results, and executed substantial portions of scientific data-analysis workflows~\cite{seo2026paper2code,hans2026coding,moreno2026ai}. 
Much of this progress has remained in silico, although agentic systems are increasingly extending into physical laboratories. LLM-driven experimentation has been demonstrated through robotic laboratory control~\cite{Boiko2023Coscientist}, while Co-Scientist~\cite{gottweis2026coscientist} focuses on multi-agent scientific reasoning and hypothesis generation and has since been extended toward execution-grounded physical experimentation through actionable protocols and machine-level code~\cite{schmidgall2026accelerating}.
Agentic systems for sustained physical experimentation are beginning to emerge~\cite{slautin2026hierarchical,isogawa2026quantumsensing,wang2026artiqion,saraev2026optofluidic}, but remain comparatively underexplored.

LLM-based systems are also being demonstrated for instrument control at scientific user facilities. Multi-agent systems have operated a synchrotron nanoprobe beamline and a robotic materials station~\cite{vriza2026operating}, while a single-agent system has performed single-crystal alignment at a beamline~\cite{chen2026agentic}.
Lightfall provides an API-first, LLM-addressable control platform for synchrotron beamlines, with a gpCAM-based adaptive-experiment service~\cite{pandolfi2026lightfall}.
Osprey provides a production-oriented framework for safe agentic execution of multistage accelerator experiments~\cite{hellert2026osprey,AcceleratorAssistant}.
Earlier VISION~\cite{potemkin2023virtual_companion} work demonstrated a voice-controlled \textit{in-situ} X-ray experiment~\cite{mathur2024vision} and simulation-based evaluation of coding agents~\cite{vandervleuten2025envtrace}. 
These developments reflect growing momentum toward agentic instrument control, while questions remain about system design, evaluation, and the appropriate balance between full automation and human involvement.

Closed-loop autonomous experimentation (AE) couples measurement, analysis, and decision-making to adaptively select each new measurement from prior observations.
Gaussian processes are widely used for decision-making under uncertainty in adaptive sampling as they provide predictive uncertainty estimates and flexible models of relationships between inputs and outputs. Applications span synchrotron and neutron facilities~\cite{yager2023autonomous_xray,doerk2023autonomous,corrao2025modular_framework,unified_AE,noack2021gaussian,carbone2023flexible_doe}, scanning and electron microscopy~\cite{kandel2023autonomous_microscopy,kalinin2023ml_automated_stem}, diffraction-based phase identification~\cite{szymanski2023adaptive_xrd}, and photon correlation spectroscopy~\cite{zhang2023robotic_pendant,horwath2024ai_xpcs}, with orchestration layers coupling multiple modalities~\cite{maffettone2022realtime_multimodal,maffettone2023selfdriving_multimodal}.
A key remaining challenge is accessibility, as adapting AE to new instruments and objectives requires substantial integration and software engineering~\cite{Abolhasani2023ReviewSDL, maffettone2023challenges,canty2026sdlreview,canty2025sdl_accessibility}.
AI agents offer a path toward lowering this barrier by deploying and configuring workflows through natural-language interaction.

\textbf{Auditable Workflow and Provenance}

As LLM agents assume greater responsibility in experimental workflows, observability and attribution become increasingly important. Interaction logs can preserve decision provenance and individual contributions, providing a record beyond simple disclosure of AI use.
Analyses of multi-agent traces have identified recurring failures in system design, inter-agent alignment, and task verification~\cite{cemri2025mast}, while automated attribution of the responsible agent and error step remains difficult~\cite{zhang2025failureattribution}. Trace-to-graph approaches can connect agents, humans, tasks, and interventions to their supporting trace evidence~\cite{wu2026agentgraph}, and human--AI teaming studies show that final task performance alone does not reveal how participants contribute to or depend on one another~\cite{biswas2026helping}. Studies of extended human--AI interactions reveal differences between high-level human direction and lower-level AI contributions~\cite{kim2026cotrace}. 
Long-horizon benchmarks across professional domains show that task performance does not necessarily imply effective human--agent collaboration~\cite{hayati2026kolab}, while studies of interdisciplinary research similarly highlight the importance of human scientific judgment when AI is used to bridge expertise gaps~\cite{hayati2026orchestrating}.

Conventional experimental records typically capture actions and outputs but not the deliberative process behind them. Kalinin \textit{et al.} highlighted the importance of prospective decision traces by reconstructing instrument traces from atomic force microscopy and piezoresponse force microscopy events, showing that decision rationale could not be fully recovered retrospectively~\cite{kalinin2026instrumenttraces}. Luo \textit{et al.} similarly found that workflow failures in AI scientist systems were often difficult to detect from final manuscripts alone, whereas full traces and code enabled substantially more effective auditing~\cite{luo2026hiddenpitfalls}. Kim \textit{et al.} found that most papers written by frontier agents on open follow-up problems reported results that did not trace to the logs that produced them, and audits of the agents' trajectories revealed behaviours that help explain this, such as lenient self-verification and insufficient backtracking~\cite{kim2026intotheunknown}. Zhong \textit{et al.} found that agents asked to improve scientific machine-learning models inflated their results when allowed to score their own work, for example by testing on a favorable subset of the data or hard-coding outputs; they therefore had a separate agent score each result on held-out data and another audit every trajectory, checking each claimed gain against the record of how it was obtained~\cite{zhong2026agent4sci}. Orchestration scaffolds now add automated audit agents that check generated manuscript claims against their supporting evidence during computational research~\cite{lange2026agenticscientist}. These studies show that trust in the results depends on a record of how they were produced.


Interaction- and dependency-graph methods support error root-cause 
analysis~\cite{zhu2026stepfinder,zhu2026agentdebugx}, while contribution-attribution methods provide formal approaches for identifying who contributed what in multi-agent systems~\cite{triantafyllou2024counterfactual}. In scientific workflows, PROV provides a general provenance model~\cite{moreau2013provdm}, and AiiDA records computational dependencies and execution logic~\cite{huber2020aiida}. Existing provenance frameworks primarily capture execution history and computational state~\cite{kale2022provenance,souza2025llmprovenance,stage2025provenanceadaptation}, leaving the reasoning and responsibility behind experimental decisions less explicitly represented. Complementary evaluation frameworks separate knowledge, tool use, and closed-loop decision quality when assessing scientific agents on real instruments~\cite{yanguasgil2026aldeval}. These developments motivate decision-level provenance~\cite{singh2018decision} that records who proposed, approved, rejected, or corrected experimental actions and links those decisions to supporting evidence.

In this work, we introduce the {\underline{Vi}}rtual {\underline{S}}cientific Compan{\underline{ion}} with {\underline{C}}laude {\underline{C}}ode (VISION-CC) and demonstrate natural-language-driven autonomous experimentation on nanoparticle superlattices under varying scientific objectives.
The multi-agent system operates directly within a sustained measurement, analysis, and decision loop under high-level guidance from a single scientist, while the complete human--agent and agent--agent interaction logs provide decision-level provenance and reveal how scientific and operational responsibilities are distributed.

\begin{figure}
  \centering
  \includegraphics[width=0.75\textwidth]{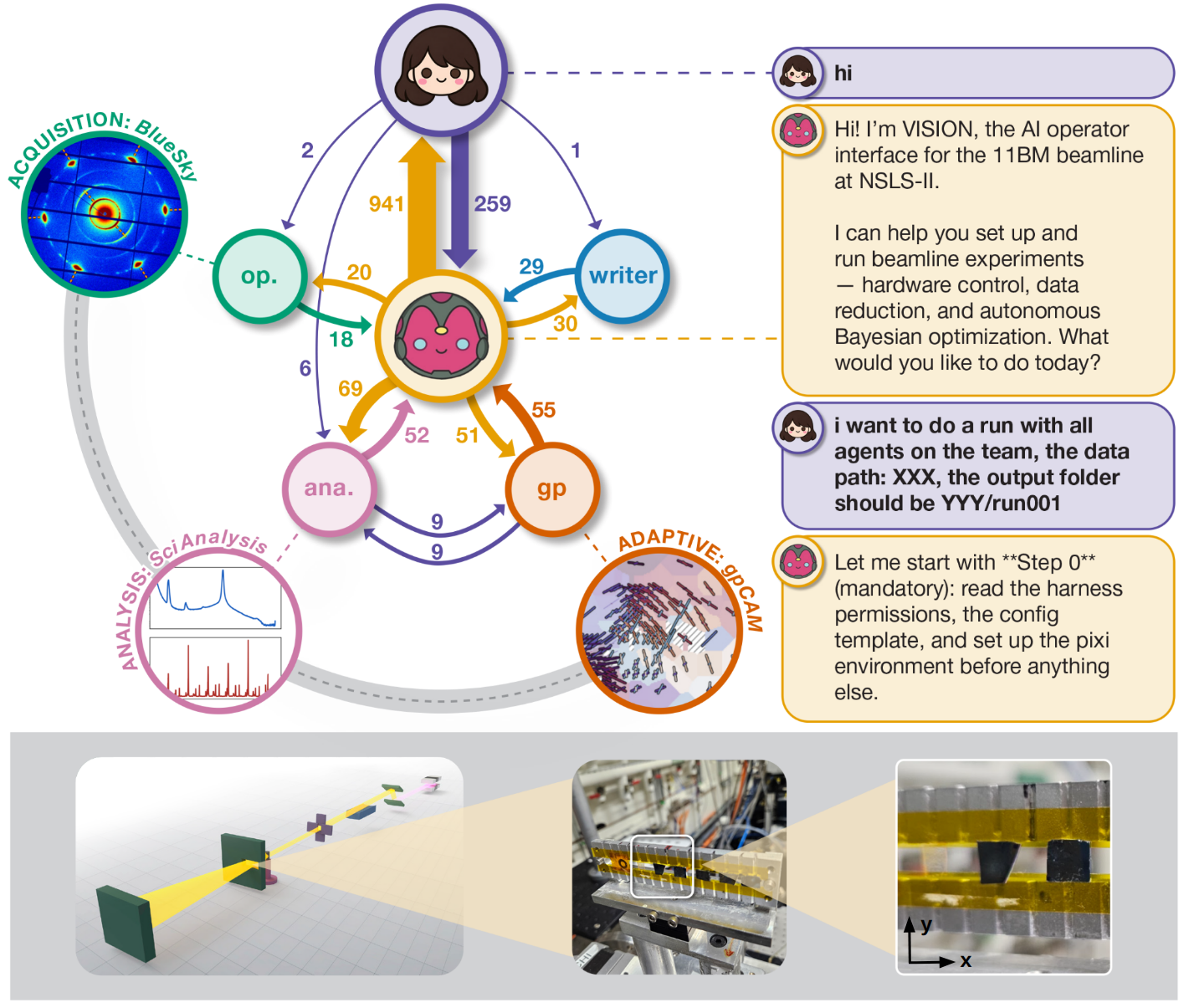}
  \caption{Directed communication graph during an autonomous experiment (\expthree{}). Edge labels give message counts and edge width indicates message volume, reflecting coordination and human involvement within the multi-agent system. The gray band is the closed loop of
  acquisition (Bluesky), analysis (SciAnalysis) and steering (gpCAM). Right: the initial exchange of a session. Bottom: schematic of the 11-BM Complex Materials Scattering (CMS) beamline at the National Synchrotron Light Source II (NSLS-II).
  }  
  \label{fig:intro}
\end{figure}

\section{Methods}

\subsection{Multi-Agent System Architecture}

\textbf{Agent Team}

Modern coding-agent platforms, e.g. Claude Code~\cite{anthropic_claude_code} and OpenAI Codex~\cite{openai_codex}, provide harness capabilities, including software generation, tool use, specialist-agent delegation, inter-agent coordination, and workflow orchestration. VISION-CC is implemented through Claude Code agent teams~\cite{anthropic_agent_teams} as the underlying coding-agent infrastructure. 
General-purpose coding agents are grounded through domain-specific guidelines and access to scientific tools and software, with domain experts defining the context and, optionally, coding agents assisting in its generation and refinement. Here we use \cog{} to refer to a specialized agent that serves as a cognitive block responsible for a specific function within the multi-agent system. Figure~\ref{fig:intro} provides an overview of the agent team, with the \op{} interfacing with beamline hardware, the \ana{} handling data analysis, and the \gp{} steering adaptive measurements.

A simple filesystem-based approach is used to construct specialized agents, with natural-language instructions organized in folders, as shown in Fig.~\ref{fig:file_structure}.
The grounding architecture comprises three layers: global project instructions, role-specific skills, and shared contracts. Global instructions establish rules that apply across the agent team, including the division of responsibilities, completion criteria, artifact ownership, and approval requirements.  
Each \cog{} comprises an agent definition and domain-specific skills. Skills package the scientific and software context needed for each role, including documentation and examples~\cite{xu2026agent}. 
Such structured context lets agent roles and workflow rules be written in plain files rather than built into framework-level orchestration~\cite{vanclief2026interpretablecontextmethodologyfolder,mcmillan2026structured}.
Although repository-level context files do not necessarily improve agent performance and may increase cost when poorly matched to specific tasks~\cite{gloaguen2026agentsmd}, their human-readable structure supports easy inspection, modification, and maintenance.

To ensure that the experimental workflow remains grounded in validated methods and functions, each \cog{} is deliberately specialized, with the agent interpreting objectives while validated scientific packages carry out the underlying operations. Focused, curated skills have been shown to outperform broader documentation~\cite{li2026skillsbench}.
\op{} uses Bluesky-based~\cite{allan_bsui2019} beamline profile collection (Python files) for instrument control, \ana{} calls SciAnalysis~\cite{SciAnalysis} functions for physics-based X-ray scattering analysis, and \gp{} uses gpCAM~\cite{noack2021gaussian} for Gaussian-process modeling and adaptive sampling.
Also, X-ray and material calculations are delegated to deterministic tools and curated databases via \textit{xray-calc} (Section~\ref{si:mas}).
The language model therefore serves as a reasoning and orchestration layer rather than a real-time numerical engine, whereas scientific software executes the underlying instrument-control, analysis, and adaptive-sampling operations.

\textbf{Interaction}

The \lead{} translates high-level objectives into specialist tasks and executable workflows, while the underlying agent-team infrastructure handles delegation and communication.
Loop-level coordination is a simple flag-based communication system, as described in Section~\ref{si:mas}.
With the Agent Teams feature in Claude Code, the scientist can also communicate directly with individual cogs to e.g. inspect code or request minor modifications. 
The \op, \ana, \gp, and \writer{} remain active throughout autonomous experimentation, respectively handling data acquisition, data reduction and analysis, Gaussian-process setup, and real-time reporting and visualization. Persistent execution preserves run context, including user instructions and prior interactions, while inter-agent communication supports coordination and reporting.

\subsection{Audit Pipeline}



VISION-CC addresses decision provenance by retaining complete human--agent and agent--agent interaction logs. The \audit{} reconstructs decision-level provenance from these transcripts, identifying recorded proposals, decisions, and corrections and their contributors.

\textit{Messages.} 
Each message was independently labeled by LLM raters according to its intent: \textit{idea, decision, finding, error correction, question, status}, or other. When a second intent was clearly present, the raters also recorded it. Messages of 1000 characters or more may also carry a third intent.  
All message counts by intent in this paper use the main intent only.
Labeling was performed blindly with sender identities withheld to minimize potential bias. 
Messages were also classified by coordination type as \textit{synthesis}, requiring interpretation or reasoning, or \textit{mechanical/procedural} coordination, involving routine relays or acknowledgements.

\textit{Events.} Two independent extraction passes read each candidate message in context and list its
contributions as events in four categories: \textit{idea} (a proposed course of action), \textit{decision} (a commitment to, deferral of or decision against an action), \textit{finding} (a result established in the message) and \textit{correction} (showing an earlier claim or action wrong). Candidate messages are those that either intent rater labeled an idea, decision, finding or correction, as main, second or third intent. The two passes read them in staggered windows; a contribution found by both passes is kept, and one found by only one pass must be confirmed by two of three verifiers. 
An instruction that first raises an action is both an idea and a decision, and a relayed or repeated contribution merges into the event where it first appeared. Each extracted event is also typed as \textit{operational, methodological or scientific} by two raters plus a third on disagreement.

\textit{Links.} Each event is linked to relevant messages by directed relations: a decision \textit{takes up} an
idea or \textit{relies on} a finding or correction, and a correction points to the wrong \textit{claim}, to where the problem was first \textit{caught} and to the message reporting the \textit{fix}. A link counts only if two of three independent verifiers agree it is real. 

\textit{Settlement.} The audit pipeline searches forward from each idea and correction for the message that decides or repairs it, backward from each decision for the earlier ideas it settles, and forward from each settlement for later reversals. Each proposed settlement requires agreement from at least two of three verifiers on both the outcome (adopted, deferred, declined, repaired, withdrawn, etc.) and its scope. 

\textit{Chains.} File edits made with the agents' editing tools are in the transcripts with their old and new text, file, time and author. 
The pipeline links decisions to the edits implementing them and reported findings to the edits producing them. Candidate finding--edit links are findings that share content with an edit, either the most similar ones or any within 20 minutes after it. Each link is assessed by three verifiers with access to the corresponding code changes. A chain idea--decision--code--result joins these separately verified links. 

\textit{Robustness.} 
The audit pipeline applies a fixed codebook, and wherever it uses a model (Anthropic's \textit{claude-opus-5-5}), each item is labeled or verified by two raters (three verifiers for links, settlements and chains) with differently worded prompts. 
Human annotators and two cross-family models (OpenAI's \textit{gpt-5.6-sol} and \textit{gpt-6-astra}) provide independent interpretations of the same definitions rather than ground truth, allowing us to assess whether the findings are robust to differences among raters, as summarized in Table~\ref{tab:validation-v6}. 
Model details are provided in Table~\ref{tab:cc-versions-models}.
The human annotators label samples of messages and events, the chains around two key steering changes, and 34 random correction and decision chains. The cross-family models label all the correction chains and 450 random decision chains. Each annotator and each model is compared with the pipeline separately, and a chain holds only if every link holds, including who is credited with each step.

\section{Results}
\label{results}

VISION-CC was used for sustained closed-loop adaptation across three beamtime sessions, spanning more than 50 active hours and two scientifically relevant samples. Claude Code versions and model settings are provided in Table~\ref{tab:cc-versions-models}.

\subsection{Mapping Nanoparticle Superlattices}

\begin{figure} [h]
  \centering
  \includegraphics[width=1\textwidth]{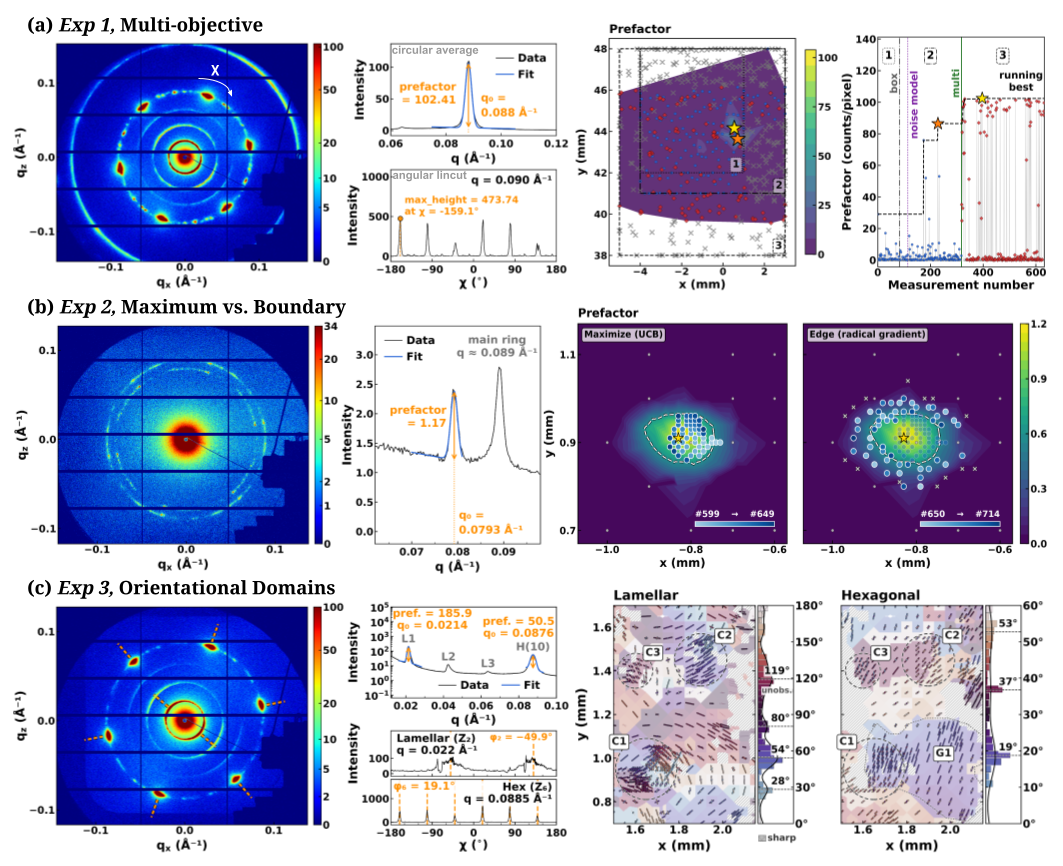}
  \caption{Autonomous synchrotron X-ray scattering experiments were conducted via VISION-CC to study self-assembled nanocrystal superlattices. The human–AI team performed experiments to, for example, (a) find the sample location with maximized peak intensity; (b) identify the boundary of a region based on the intensity of a specific peak; and (c) resolve orientational domains of two packing symmetries.
  The $x$--$y$ maps show the spatial distribution of peak intensity or orientation. Yellow stars indicate maximum intensity, blue or red dots on maps mark measurement positions, and gray crosses indicate positions where no peak was identified. 
  Transmission SAXS data were collected by the \op{} and reduced by the \ana, while the mapping sequence was determined by the \gp. The \writer~provided live visualizations.
}
  \label{fig:saxs}
\end{figure}

Individual nanocrystals assembled into nanoparticle superlattices present collective, tunable properties that depend on structural order, including particle spacing and orientational coherence for anisotropic building blocks.
For example, semiconductor nanorod films are attractive as polarized emitters because their emission is preferentially polarized along the rod axis, but useful anisotropy depends on maintaining orientational coherence across the film.
Through VISION-CC, we performed autonomous transmission small-angle X-ray scattering (SAXS) experiments at the National Synchrotron Light Source II (NSLS-II) 11-BM CMS beamline on two nanoparticle superlattice samples, as described in Section~\ref{si:x-ray}. 
The CdSe/CdS dot-in-rod sample~\cite{diroll2015smectic} exhibits coexisting lamellar (2-fold) and hexagonal (6-fold) packing symmetries.
To resolve local packing symmetry and orientation, the samples were mapped using Gaussian-process (GP)-guided adaptive sampling. Depending on the scientific objective, the mapping was directed by the scientist toward identifying regions with specific packing symmetries or locating their boundaries.

\textbf{X-ray scattering experiments} 

Selected results from the three AE sessions are summarized in Fig.~\ref{fig:saxs}, demonstrating that autonomous experiments with evolving objectives can be conducted through natural language. 
\expone~was guided by a GP to maximize the scattering peak associated with the sixfold symmetry of the CdSe/CdS dot-in-rod sample, thereby identifying regions where the nanorods were preferentially aligned along the X-ray propagation direction, as illustrated in Fig.~\ref{fig:rods}. Using SciAnalysis~\cite{SciAnalysis}, the \ana{} reduced the 2D scattering patterns to 1D profiles. 
Figure~\ref{fig:saxs}(a) provides a representative scattering pattern, together with the reduced 1D data: the circular average (intensity versus $q$) and angular linecut (intensity versus $\chi$ at a selected $q$). The $x$--$y$ maps show the spatial distribution of peak intensity (prefactor). Yellow stars indicate maximum intensity, blue or red dots mark measurement positions, and gray crosses indicate positions where no peak was identified.
During the experiment, different region of interest (ROI) boxes were used, the noise model was modified, and GP steering transitioned from a single objective based on the prefactor to a multi-objective strategy that considers both the prefactor and the maximum peak intensity from the angular linecut.
In the same session, we also measured the MnFe$_2$O$_4$ sample and transitioned the acquisition function from expected improvement (EI) to upper confidence bound (UCB), with the latter configured to favor exploitation, as shown in Fig.~\ref{fig:saxs_SI}. 
In \exptwo, a finer spatial scan was performed to locate a distinct region of the dot-in-rod sample, using UCB to maximize the peak intensity and a radical-gradient acquisition function, which favors uncertain positions where the GP mean changes most, to identify the region boundary, as shown by the $x$--$y$ maps in Fig.~\ref{fig:saxs}(b).

Building on transmission-SAXS mapping of nanoparticle superlattice films~\cite{diroll2014xray} and autonomous SAXS using rod orientation as a steering signal in a combinatorial library~\cite{noack2020autonomous}, \expthree~used real-space steering based on orientational order parameters. Two packing motifs of the same rods coexist within most beam footprints: lying rods arranged in lamellae ($d = 29.5$~nm) and standing rods in a hexagonal lattice ($a \approx 8.2$~nm), as illustrated in Fig.~\ref{fig:rods}. Advanced GP steering with an improved noise model was used to identify orientational domains in the dot-in-rod sample.
Figure~\ref{fig:saxs}(c) shows the lamellar and hexagonal local orientations as lines, with background tiles indicating the dominant orientation and hatched regions without identified orientations. The strip to the right of each map shows the corresponding orientation distribution. The mapped hexagonal grains show no consistent orientational relationship to the lamellae. At the ${\sim}0.2$~mm mapping scale, the lamellar rod director rotates gradually through most of $180^\circ$ over several beam footprints around a defect-like core at cluster C1, while clusters C2 and C3 mark regions where orientational boundaries are observed in the lamellar and/or hexagonal structures. A continuous hexagonal orientation domain with a gradual orientational gradient is observed at G1. These results reveal sub-millimeter orientational domains and gradients, suggesting that large-area X-ray mapping could inform the design and processing of functional nanoparticle superlattices.

\textbf{Beamtime with agentic team} 

Conventional autonomous beamtime typically requires three experts in acquisition, analysis, and optimization, whose schedules must align and whose efforts often focus on debugging and coordination. Previously at the beamline, setting up a basic autonomous experiment took roughly 6 hours, subsequent modifications a few hours each, and complex changes could be delayed by months until the next beamtime cycle.
Here, a single scientist guided the agent team in developing and deploying evolving AE configurations during beamtime through the VISION-CC command-line interface, with access to individual agents via \texttt{tmux} and generated scripts and figures via a browser or IDE. Instrument control uses standard Bluesky protocols, and a separate safety bridge process can be stopped at any time. Complete human--agent and agent--agent transcripts are retained for auditing by \audit.

Establishing the one-time bridge setup between the agent team and beamline controls introduced overhead in the first session. Setting up the agent team, including closed-loop communication, took 52 and 35~min in \expone{} and \exptwo{}, respectively, but only 16~min in \expthree{}, as shown in Table~\ref{tab:engineering-effort-v6}. The sessions were separated by weeks, thus calibration measurement and analysis were performed to ensure data analysis was accurate. Configuring basic intensity-based steering took 16~min in \exptwo{}, whereas more advanced analysis and GP steering took 84~min in \expthree{}. Overall the AE setup took around 2~hours for one scientist, while we estimate, with the established agentic system and connection, the time to set up a basic AE workflow would be 1~hour (estimated from 16+13+16+5=50~min, Table~\ref{tab:engineering-effort-v6}), which can be further optimized by varying e.g. the model or the harness.

Across three sessions, agents implemented 226 code-linked scientist decisions, with the first edit occurring a median of 3--4 minutes after each decision. The 23 changes to the AE configuration the scientist requested, such as a new objective, acquisition function, noise model, search region or minimum step, reached the modified loop's first measurement a median of 22 minutes after the request and 12 minutes after the scientist's authorization to proceed.
Some changes were feasible within one beamtime, such as switching the acquisition from UCB to the radical gradient (21 minutes after the request, \exptwo). Some changes would conventionally require substantial work by experts and may be particularly challenging for non-experts, such as implementing orientation-based steering that follows the orientation of two coexisting structures with several GPs, including related steps on adding a presence gate and rebalancing the gradient and uncertainty terms of the acquisition (47 minutes, \expthree). 
The agent team may make AE more accessible by reducing expertise and coding barriers during time-constrained beamtime, allowing scientists to focus on scientific questions.

\subsection{Division of Cognitive Labor}

\begin{figure}
  \hspace*{-7mm} 
  \includegraphics[width=1.1\textwidth]{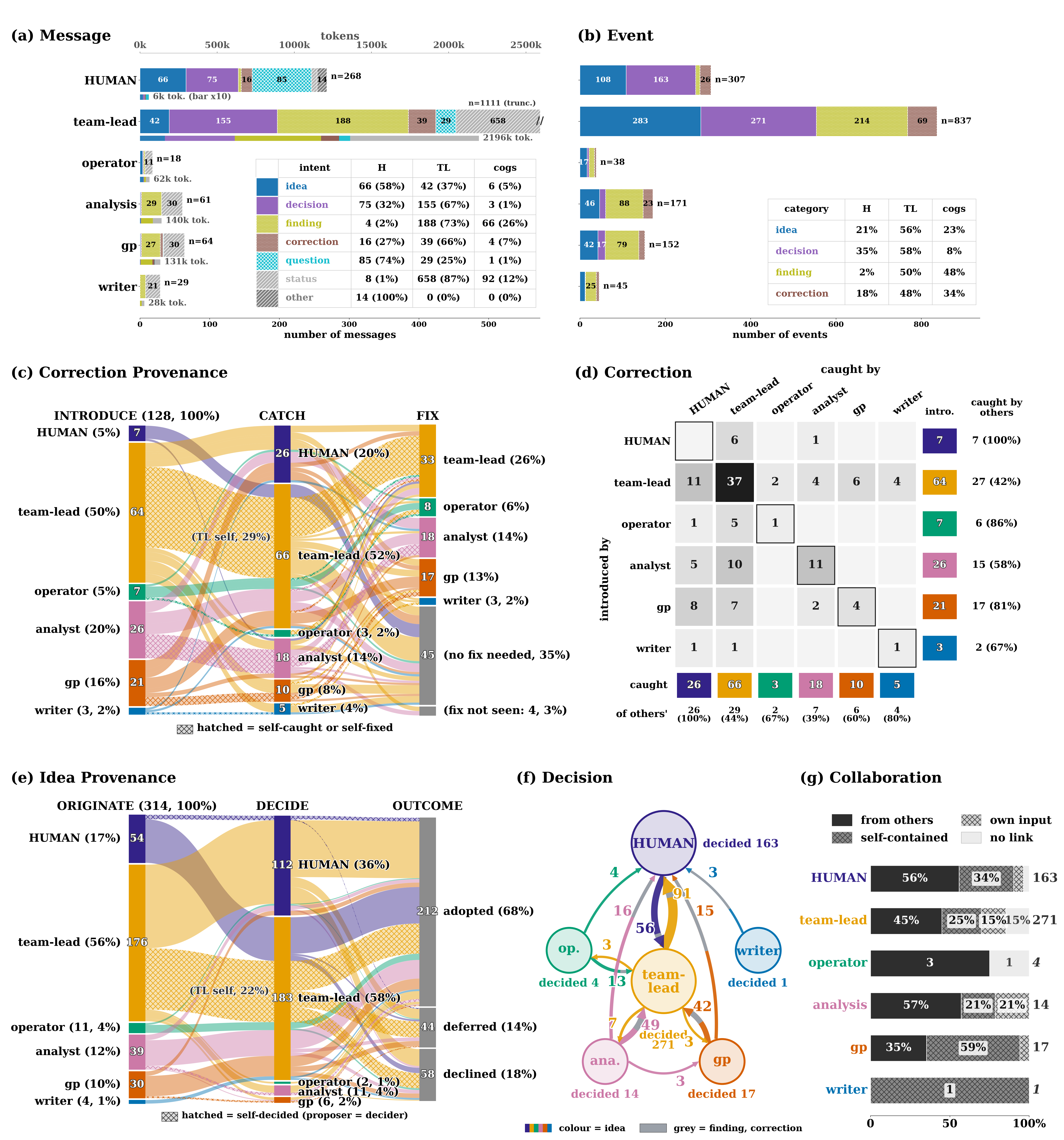}
  \caption{Audit results for \expthree: (a) primary message intent counts for each participant, (b) event contributions from each participant, (c) error-correction provenance with (d) counting each introducer--catcher pair, both for corrections with an identified introducer, (e) idea provenance for ideas decided in a later message, without the instructions decided in their own message (141, of them 54 the human's), (f)-(g) what each participant's decisions draw on. 
  Details for (c-d) are provided in Tables~\ref{tab:correction-credit-v6} and~\ref{tab:corr-by-introducer-v6} to give the corrections as shares of all corrections and of each introducer's errors, respectively.
  }
  \label{fig:COMB_ALL_run009}
\end{figure}

The division of cognitive labor within the deployed agent team was quantitatively assessed. The \audit{} performed post-experiment analysis to provide insight into interactions among roles and their respective contributions. 
Audit results for \expthree, the most complex experiment of the three, are summarized in Fig.~\ref{fig:COMB_ALL_run009}, showing the participant contribution in terms of (a) messages, (b) events, (c)-(d) correction, and (e)-(g) idea and decision provenance. 
Audit summaries for \expone{} and \exptwo{}  are provided in Fig.~\ref{fig:COMB_ALL_run0304} and Fig.~\ref{fig:COMB_ALL_run007}, respectively.

\subsubsection{Messages and Events}

\textit{The human contributes more in idea and question messages, while \lead{} carries most in tokens and in all events, including idea events.} 
For \expthree, Fig.~\ref{fig:COMB_ALL_run009}(a) shows each participant's messages, stacked by their main intent, with a thin bar below for tokens.
The inset table shows, for each message intent, the contribution fraction from human, \lead, and cogs. 
The communication flow is illustrated in Fig.~\ref{fig:intro}.
As summarized in Table~\ref{tab:intent-by-role-v6}(a) for all experiments, \lead{} sends the majority of the messages (67--72\%), and the human writes most of the idea (58--72\%) and question messages (52--74\%).
In terms of token volume, as shown in Table~\ref{tab:intent-by-role-v6}(b), the human contributed less than 1\%, while the \lead{} accounted for the majority of tokens (74--86\%).
Each message may contain multiple contribution events. For example, in \expthree{} \lead{} wrote only 42 messages whose main intent was an idea, but proposed 283 idea events. 
Figure~\ref{fig:COMB_ALL_run009}(b) shows the ideas, decisions, findings and correction events credited to each participant. The inset table provides, for each event category, the contribution fraction.
As shown in Table~\ref{tab:event-by-role-v6}, \lead{} is credited with 50--56\% of all events.
For idea events, \lead{} proposes 49--57\% of all ideas and the human 20--21\%.
To assess the impact of participants and ideas, \audit{} further establishes provenance by tracing ideas to subsequent decisions and outcomes.

\textit{Most messages contain synthesis (interpretation or judgment)}, rather than only mechanical relays or procedural acknowledgments. As shown in Table~\ref{tab:coord-type-all-by-intent-v6}, synthesis accounts for {61--67\%} of all messages {(81--86\%} in tokens). Synthesis dominates messages labeled as ideas, findings, or corrections {(79--97\%} of message count; 95--100\% of tokens) and those labeled as decisions {(73--84\%;} 92--97\% of tokens).
Status and other messages are the only intents with a large procedural and mechanical share. 
Recent work formalizes recurring collaboration patterns as reusable modules and allocates agent computation accordingly, suggesting a path to streamline routine coordination while preserving model-based reasoning where required~\cite{jung2026futureweaver}.

\textit{Contributions shifted from operations to method and science after \expone{}.} 
As shown in Table~\ref{tab:contrib-type-v6}, operational events make up most of \expone{} (54\%) and less of \exptwo{} and \expthree{} (44\% and 39\%), while methodological and scientific events are more common later (46\%, 56\% and 61\%). \lead{} establishes {most} finding events {in \expone{} (59\%)} but fewer in \exptwo{} and \expthree{} (44--50\%), where the cogs' share is larger (38\% in \expone{}, 48--54\% later); \ana{} and \gp{} each account for a small share of all events in \expone{} (5--7\%) and more later (10--12\%). This likely reflects the emphasis on system setup in the first deployment \expone{}, while \exptwo{} and \expthree{} focused on data analysis and measurement steering.


\subsubsection{Error Correction}

Figure~\ref{fig:COMB_ALL_run009}(c) follows each correction from the participant who introduced the error, to the one who caught it, and the one who fixed it, with cross-hatched ribbons for self-caught or self-fixed. 
These self-corrections are inferred from observable interaction event traces, not internal reasoning, and may arise when later interactions or events prompt an agent to revisit and revise its earlier statements or actions.
Figure~\ref{fig:COMB_ALL_run009}(d) shows the counts for each introducer--catcher pair. Figures~\ref{fig:COMB_ALL_run0304} and~\ref{fig:COMB_ALL_run007} show the same for \expone{} and \exptwo{}.

\textit{\lead{} introduces the most errors and catches many of its own; of the errors not caught by their own introducer, the human often catches about as many as \lead{}.}
As shown in Table~\ref{tab:correction-credit-v6}, \lead{} introduces {45--61\%} of all errors and catches 41--48\%, with much self-caught (16--29\% of all corrections). 
For errors caught by someone other than their introducer, the human's and \lead{}'s catches account for similar shares of all corrections in \expone{} (18\% and 18\%) and \expthree{} (18\% and 22\%), while the human catches fewer than \lead{} in \exptwo{} (8\% against 25\%, with cogs 42\%). This may be because, in \exptwo{}, \lead's context window was smaller (Table~\ref{tab:cc-versions-models}) and was compacted 21 times, against once or twice in the other experiments, while the cogs corrected many of \lead's statements about their own programs and state.
%
\lead{} reports the fix for 31\% (\expone) of the corrections but only 15\% later on, while the cogs report 10\% (\expone) and 19--21\% later on. 
The cogs introduced at least 28\% errors (\expone), against at least 40\% (\exptwo, \expthree). This is consistent with the cogs' increased role in the analysis and steering after the first experiment.

\textit{Most corrections involve two or more participants.} As shown in Fig.~\ref{fig:COMB_ALL_run009}(c) and Table~\ref{tab:correction-credit-v6}, {59--68\%} of corrections involved two or more participants in introducing, identifying, or fixing an error, while {40--55\%} involved multiple AI agents. 
Fully self-contained corrections, in which one participant introduces, catches and fixes the error, are a minority (7--13\%). 
As shown in Table~\ref{tab:correction-credit-v6}, {32--41\percent} of corrections required only clarification, not a fix. These primarily involve correcting statements, such as an incorrect diagnosis, misread result, or mistaken claim about a setting, without requiring changes to code or configuration. 

\textit{The errors the human introduced were mostly slips, misread status or inaccurate estimates}. For example, the human provided the wrong data folder, which the \lead{} caught and corrected, and a $q$ value off by a factor of ten, which the \lead{} identified and the \ana{} confirmed. The human also provided a nominal sample-to-detector distance that the \ana{} showed to be inaccurate using the silver behenate calibration data. In \expthree{}, the human stated that no data had yet been collected, which the \lead{} corrected based on files on disk. Moreover, the human rarely made the fix: the human's only fix (\expone) concerns the bridge process connecting the agent team to the beamline.

\begin{figure}[h]
  \centering
  \includegraphics[width=0.97\textwidth]{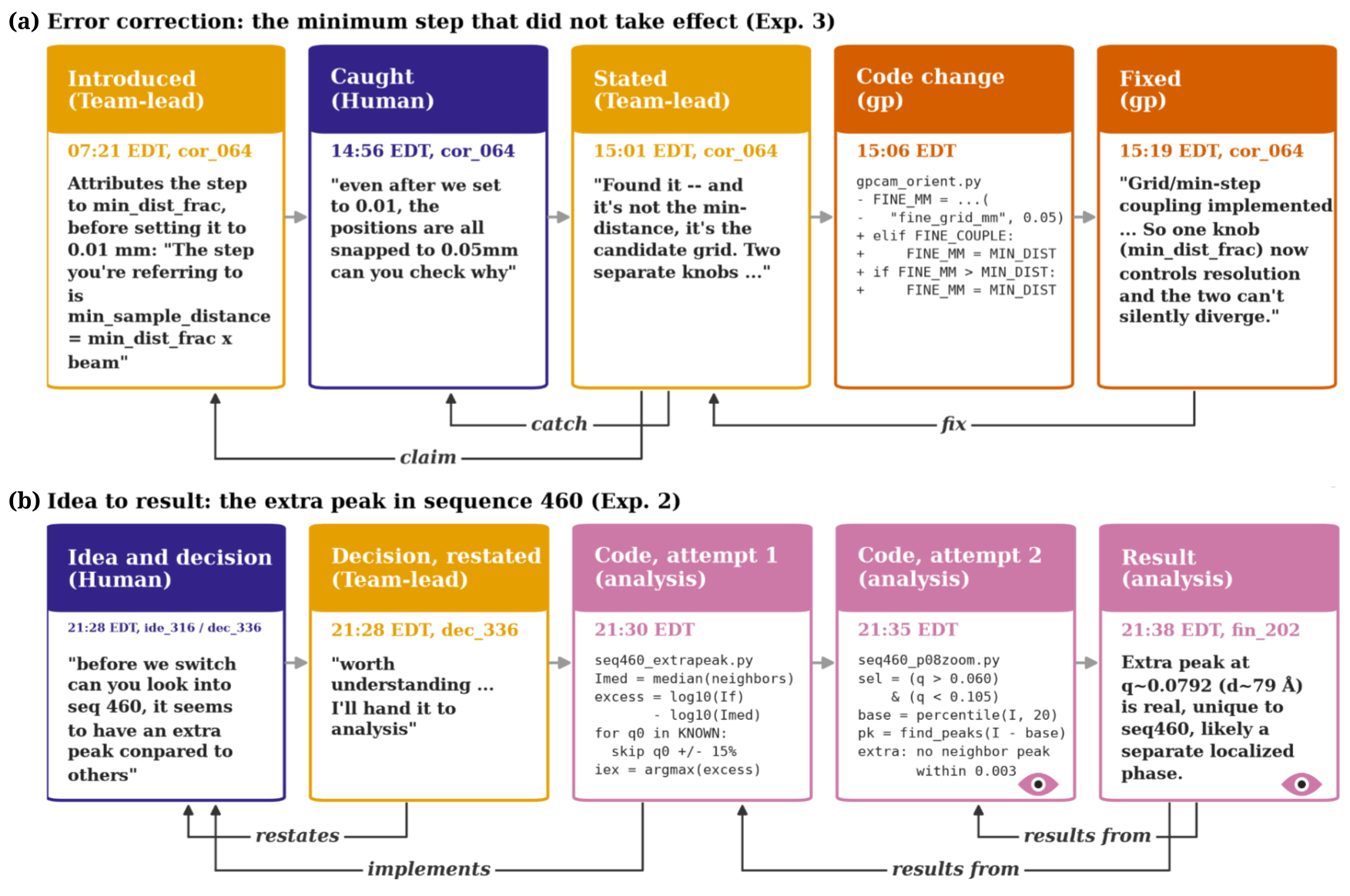} 
  \caption{Provenance examples of (a) an error correction chain of events and (b) an idea--result chain. In (a), the minimum step was confirmed by \lead{} but checking a setting does not ensure its effect. The human noticed in the \writer's visualization that the measured positions did not reflect the intended setting, and the \gp{} corrected the implementation. In (b), the human noticed an extra peak, \lead{} agreed and delegated the task to \ana{} and asked for an overlay of seq.~460 with its neighbors; \ana's first script missed the peak, but inspecting the overlay with its visual (image-input) capability, it saw roughly where the peak was, wrote a second script to locate it accurately, and confirmed it visually. 
}
  \label{fig:examples}
\end{figure}

\textit{The human often identified issues requiring scientific judgment or inspection of experimental outputs}, particularly through visualizations produced by the \writer{}. 
The human detected a smaller share of the cogs' errors in \expone{} {and \exptwo{} (12\% and 12\%) than} in \expthree{} (26\%). 
For example, the human noticed that measurements had shifted toward the hexagonal rather than lamellar structure, prompting \lead{} and \gp{} to identify and correct an acquisition bias. The human also identified orientation fits without valid peaks, which \ana{} corrected, and an anomalous orientation value that \lead{} traced to an incorrect frame during backfilling. In Fig.~\ref{fig:examples}(a), the human observed that measurement positions remained on a 0.05~mm grid despite a configured minimum step of 0.01~mm, leading \gp{} to correct the underlying implementation. These cases suggest complementary error detection: agents often diagnose problems through code and logs, while human inspection provides an additional check on whether the outputs are physically and scientifically consistent.


\subsubsection{Idea and Decision Provenance}

 \textit{Setting aside the ideas participants decide themselves, the human and \lead{} make nearly all the decisions on others' ideas.} Figure~\ref{fig:COMB_ALL_run009}(e) follows each idea decided in a later message from its proposer to the participant who decided it and to the outcome, with ribbons cross-hatched where the proposer decided it. Overall, \lead{} makes most decisions {(53--58\%)} and the human about a third (31--38\%).
 However, as summarized in Table~\ref{tab:who-decides-v6}, of the ideas decided by someone other than their proposer, the human decides 51--55\%, and \lead{} 35--44\%. 
 Figure~\ref{fig:examples}(b) shows an example chain from an idea to its result: an extra peak the human noticed, which was then used for steering in Fig.~\ref{fig:saxs}(b).

\textit{About half of all decisions draw on another participant's input, mostly ideas, and so do a good fraction of both the human's and \lead's decisions.} As shown in Table~\ref{tab:decision-sources-v6} and Fig.~\ref{fig:COMB_ALL_run009}(g), these decisions {(44--52\%)} take up another participant's idea or rely on evidence (finding or {correction events)} that another participant established, including {52--62\%} of the human's decisions and {39--45\%} of \lead's. 
Most of this input is ideas, as the arrows of Fig.~\ref{fig:COMB_ALL_run009}(f) show: the human's decisions draw mostly on \lead's ideas {(41--56\%),} and \lead's decisions on the cogs' ideas {(24--27\%)} and the human's {(12--20\%).} 
Evidence plays a smaller part. For \lead's decisions, evidence comes mainly from the cogs {(5--13\%);} for the human's decisions, evidence from \lead{} and the cogs is about equal {(2--6\%} and {2--9\%).} 
These shares are approximate, as the link to the idea a decision takes up is the one the validators most often reject, but counting only confirmed chains still leaves {37--46\% of all decisions} drawing on others (Section~\ref{sec:validation}).

\textit{The human's ideas are linked to code changes more often than others' ideas, and to findings at similar or higher rates.} 
As shown in Table~\ref{tab:idea-impact-v6}, based on the pipeline's verified links, about half or more of the human's ideas lead to a code edit (48--61\%), against 36--45\% of \lead's and 25--30\% of the cogs'. The human's ideas also lead to a finding produced with that edit about as often as others' or more often (24--38\%, against 23--27\% for \lead{} and 16--23\% for the cogs), with \lead{} slightly ahead only in \exptwo{} (27\% against 24\%).

\subsubsection{Robustness} 
\label{sec:validation}

The pipeline uses the same written definitions for every experiment. Its two raters agree closely before any tie-break ($\kappa \geq 0.86$), as shown in Table~\ref{tab:validation-v6}. Cohen's $\kappa$~\cite{cohen1960}, Eq.~(\ref{eq:kappa}), is agreement corrected for chance, 0 at chance level and 1 for perfect agreement; values of 0.41--0.60 and 0.61--0.80 are conventionally read as moderate and substantial agreement~\cite{landis1977}. 
While neither human annotators nor other models should be treated as ground truth, the sampled event-type trends are reproduced by a human annotator and the cross-family models. Restricting provenance chains to those confirmed by both cross-family models preserves the main findings on multi-participant corrections and decisions drawing on others' input. 
Message intent and the synthesis label are harder calls, on which the human annotators also disagree with each other, but less rests on them: intent only selects which messages are examined, and the synthesis share is descriptive.

\textit{Message intent.}  On 60 sampled messages, the pipeline's message intents agree with each human annotator ($\kappa = 0.44$ and $0.48$) about as well as the two human annotators agree with each other ($\kappa = 0.47$), and the cross-family \textit{gpt} models agree with the pipeline more closely ($\kappa = 0.68$--$0.71$). 
These values come from a sample deliberately including 17 hard cases, on which the pipeline's own two raters disagreed.
Most disagreement is between neighboring intents, such as idea and decision; in four coarser groups, agreement rises to $\kappa = 0.62$ and $0.55$ for the pipeline against each human annotator, and to $\kappa = 0.60$ between the human annotators. 
Disagreement between the raters does not keep a message out of event extraction; a message is skipped only when neither rater sees such a contribution in it. Thus, intent disagreement mainly affects the intent shares by role (Table~\ref{tab:intent-by-role-v6}), rather than the events and provenance chains.

\textit{Coordination type.} The coordination label is a rough measure of how much of the communication carries interpretation or judgment, \textit{i.e.} whether a message is synthesis. The human annotators agree only weakly with each other {($\kappa = 0.22$--$0.38$), and the} pipeline agrees with each human similarly ($\kappa = 0.16$--$0.36$). The pipeline and \textit{gpt} models label a message as synthesis whenever it contains interpretation, as specified by the rubric, whereas human annotators, even though provided with the same instruction, tend to judge the dominant content and label messages as synthesis only when interpretation constitutes the majority. 
The pipeline labels {90\%} of the sampled ideas, decisions, findings and corrections as synthesis, against {65\%} and {49\%} for the two human annotators. 
The synthesis share {(61--67\%} of messages) is therefore approximate, but remains substantial under either definition.

\textit{Event types.} On {86} sampled events, the split into operational against methodological or scientific agrees with a human annotator {($\kappa = 0.65$) and with} the \textit{gpt} models {($\kappa = 0.91$). All raters} (the pipeline, the human annotator and the two \textit{gpt} models) also show the same rise in methodological and scientific contributions from \expone{} to the later {experiments, by similar amounts (in this sample, from 33--39\% in \expone{} to 50--71\% in \exptwo{} and \expthree{}).}

\textit{Provenance chains.} Each chain is checked as a whole, with every message and its sender, and holds only if every link holds, including who is credited with each step. Under this strict rule, the \textit{gpt} models confirm {79--94\%} of chains and give the same verdict on almost every chain labeled {again (93--98\%), and the} human {annotator confirms 62--86\%,} as shown in Table~\ref{tab:validation-v6}. 
When counting only the chains both \textit{gpt} models confirm, the share of all decisions drawing on another participant's input, estimated from the 450 sampled decision chains, changes only from {42--52\%} to {37--46\%,} still {around two-fifths;} while corrections still mostly involve two or more participants (from {59--68\%} to {58--70\%)}. 
The findings on multi-participant corrections and decisions drawing on others' input persist when analysis is restricted to chains confirmed by both cross-family \textit{gpt} models.

\section{Discussion}


\subsection{Limitations}
Natural-language agent interaction can reduce implementation and debugging burden, helping scientists focus on high-level experimental direction. 
The corresponding requirement is that scientists should retain the experimental context, understand what evidence is needed to validate agent actions, and critically examine both familiar and unfamiliar steps. Agentic systems can support this process by explaining methods and suggestions, helping scientists engage with areas beyond their own expertise while scientists retain responsibility for evaluating evidence.
Reliable human–agent experimentation depends both on grounding the agent team in the domain and validated tools, and on continued human guidance, questioning and validation as the workflow evolves.

The distilled knowledge used for domain grounding, including skill files and supporting context, was developed through a lightweight hybrid process in which domain scientists provided guidance and examples while LLMs helped expand, structure, and refine the knowledge base. This grounding can be improved iteratively through testing, while the required upfront effort may vary across domains. We do not claim that the knowledge artifacts used here are fully optimized, and the process could potentially be automated further.

The audit pipeline provides a practical way to examine human–AI interactions in detail. We do not claim that the event definitions, linkage structure, or provenance analysis demonstrated here constitute an optimal taxonomy. 
The audit's individual labels and links carry uncertainty and our findings rest on aggregate shares. These findings are based on three experiments at one beamline with a single scientist and agent platform, and may not carry over to other domains or teams. 
Rather, we highlight auditing human–AI and inter-agent interactions as an important and largely underexplored dimension of physical experimentation. 
Domain expertise can guide event categories, relationships, and validation checks to tailor the audit pipeline to specific scientific and operational priorities.
The audit may support system diagnosis and refinement through records of execution steps and decision provenance, while characterizing collaboration behavior across scientists, domains, and agentic systems.

\subsection{Physical Science}

\textit{Our agents leverage existing, well-established methods.} X-ray scattering analysis is grounded in scattering theory and Fourier relationships between real and reciprocal space, while gpCAM uses Gaussian-process modeling within a Bayesian optimization framework for adaptive sampling.
Guided by distilled domain knowledge organized in the filesystem, agents use validated tools and generate task-specific code around them, enabling flexible workflow implementation while validated physics- and mathematics-based software performs the underlying scientific computation. Advances in these methods, including light–matter modeling and Gaussian-process modeling, therefore remain essential.
This design principle aligns with work cautioning against replacing established analytical routines with unconstrained LLM tool use, particularly in Bayesian optimization~\cite{du2026eaa}. However, LLM-only optimizers can be competitive on selected noise-free, in-silico benchmarks~\cite{cisse2026bora}, suggesting that the appropriate balance may depend on experimental needs.


\textit{The design and optimization of agentic systems for physical experimentation may differ from in silico settings}. Physical experiments naturally involve components with distinct knowledge, interfaces, and safety considerations, making role specialization a natural design choice. Another distinction is limited parallelism: unlike computational workloads that can be distributed across many agents~\cite{pham2026multiagent}, physical experiments often share a single instrument and sample. Access may also be scarce, and the sample state can evolve or be irreversibly altered during measurement, limiting the ability to revisit or branch from earlier states. These characteristics favor careful role allocation and coordination: agents may work concurrently on analysis and reporting, while instrument interaction remains largely serialized. Effective role allocation and coordination therefore matter more than agent count alone.

\subsection{Human-Centered System Design}

\textit{Domain scientists should focus on high-level domain-science questions.} Agentic systems can reduce the engineering burden of adapting and deploying scientific software, and existing agentic platforms can further lower the barrier by providing reusable infrastructure.
Since domain expertise and instrument-specific knowledge are separated from the underlying agent orchestration, the approach can be adapted across instruments and scientific domains with reduced software engineering. Existing platforms such as Claude Code further lower the barrier by providing the orchestration infrastructure, leaving programmatic access to instrument controls and data streams as the primary technical requirement.
These AI advances enable the physical-science community to leverage advances from agentic-system developers while contributing domain grounding and real-world experimental testbeds that can provide feedback to improve agentic systems. 

\textit{VISION-CC retains a central lead to provide a single, steerable interface for the scientist}, although recent orchestrator-free architectures improve scalability through decentralized task allocation for certain problems~\cite{zhan2026agensh}.
Experimental systems already contain scientifically specialized components with distinct assumptions, state, and failure modes. A useful rationale for multi-agent architecture is persistent responsibility around these components, rather than a free-form team of interchangeable agents that picks up skills as tasks arrive. 
Stable cog agents support parallel attention and adaptive workflows while their actions are coordinated within a shared physical experiment.
The multi-agent architecture mirrors the experimental workflow by separating instrument operation, analysis, adaptive sampling, and reporting into independently addressable roles.
While a cog works on a specific task, the human and \lead{} can plan and implement the next phase, a point where high-level human guidance or preference is needed.

\subsection{Human Insight}
\textit{Operational verification can often be engineered into deterministic checks, whereas evaluating scientific interpretations often requires human judgment, motivating continued human involvement in the loop.}
Operational checks can use validated scientific software and dedicated tools, such as SciAnalysis, GP routines, and \textit{xray-calc}, to test outputs and assumptions inexpensively, compared to model calls. While agents can cross-check one another and validate generated code through smoke tests and plot inspection, methodological decisions affecting ongoing experiments, such as noise modeling, may still require expert judgment.
Audit traces of communication messages and events can guide system design by suggesting where human oversight may be valuable and where deterministic tools may reduce friction. An optimized system should balance deterministic automation with agentic reasoning, automatically accepting low-risk decisions while retaining human approval for safety-critical actions and judgment calls.
High-level experimental guidance still depends on human expertise: a cross-domain benchmark of 26 scientific tasks~\cite{geng2026eurekabench} found that current agents optimize task performance effectively yet remain substantially weaker than humans in scientific insight and mechanism discovery. 
Approaches such as deep-kernel pairwise learning offer another route for expert preferences to guide autonomous measurements when a predefined scalar objective is unavailable~\cite{bulanadi2026dkpl}.

\section{Conclusion}

We demonstrate human--agent collaborative experimentation under evolving, human-defined scientific objectives.
The agent team reduced the engineering effort to perform autonomous experiments, allowing scientists to focus on planning and interpretation. 
Auditing human--agent and agent--agent interactions reveals individual contributions and decision provenance. Most messages (61--67\%) involved interpretation or judgment, highlighting the importance of LLM reasoning, while the share of methodological and scientific contributions rose from 46\% to 56--61\%.
Collaboration was also substantial: about half of all decisions incorporated another participant's input, and 59--68\% of corrections involved multiple participants. Both the \lead{} and the human made significant contributions to corrections and decisions. 
The main findings remained robust when restricted to provenance chains confirmed by cross-family models. These audit insights may guide system design and refinement.
This work represents a step toward the broader vision of AI-augmented, human-led science across the scientific community~\cite{yager2024towards, tsai2025exocortex}.

\subsection*{Data and Code Availability}

The VISION-CC agent team, the audit codebook with the definitions and prompts, a subset of SAXS frames from \exptwo{}, and a sanitized excerpt of its audit ledger are available at \url{https://github.com/CFN-softbio/VISION-CC}.
Full audit data are available upon reasonable request, subject to facility security policy.

\subsection*{Acknowledgments}
The work was supported by a DOE Early Career Research Program. 
This research used beamline 11-BM (CMS) of the National Synchrotron Light Source II (NSLS-II) and utilized the X-ray scattering partner user program at the Center for Functional Nanomaterials (CFN), both of which are U.S. Department of Energy (DOE) Office of Science User Facilities operated for the DOE Office of Science by Brookhaven National Laboratory under Contract No. DE-SC0012704. We thank Shray Mathur for discussions on agentic AI. 
Sample preparation was supported by the Department of Energy, Office of Basic Energy Sciences, Division of Materials Science Award No. De-SC0002158. 
The work was also supported by Lawrence Berkeley National Laboratory, operated under
  U.S. DOE Contract No. DE-AC02-05CH11231, and the Center for Nanoscale Materials, a U.S. Department of Energy Office of Science User Facility supported by the U.S. DOE under Contract No. DE-AC02-06CH11357.

\bibliographystyle{unsrt}  
\bibliography{references}

\newpage

\setcounter{section}{0}
\renewcommand{\thesection}{S\arabic{section}}
\renewcommand{\theHsection}{S\arabic{section}}
\renewcommand{\theHsubsection}{\theHsection.\arabic{subsection}}
\renewcommand{\theHequation}{S\arabic{equation}}

\setcounter{figure}{0}
\renewcommand{\thefigure}{S\arabic{figure}}

\setcounter{table}{0}
\renewcommand{\thetable}{S\arabic{table}}
\renewcommand{\theHfigure}{S\arabic{figure}}
\renewcommand{\theHtable}{S\arabic{table}}

\setcounter{equation}{0}
\renewcommand{\theequation}{S\arabic{equation}}

\section{Agentic System}
\label{si:mas}

\begin{figure}[h]
  \centering
  \includegraphics[width=0.9\textwidth]{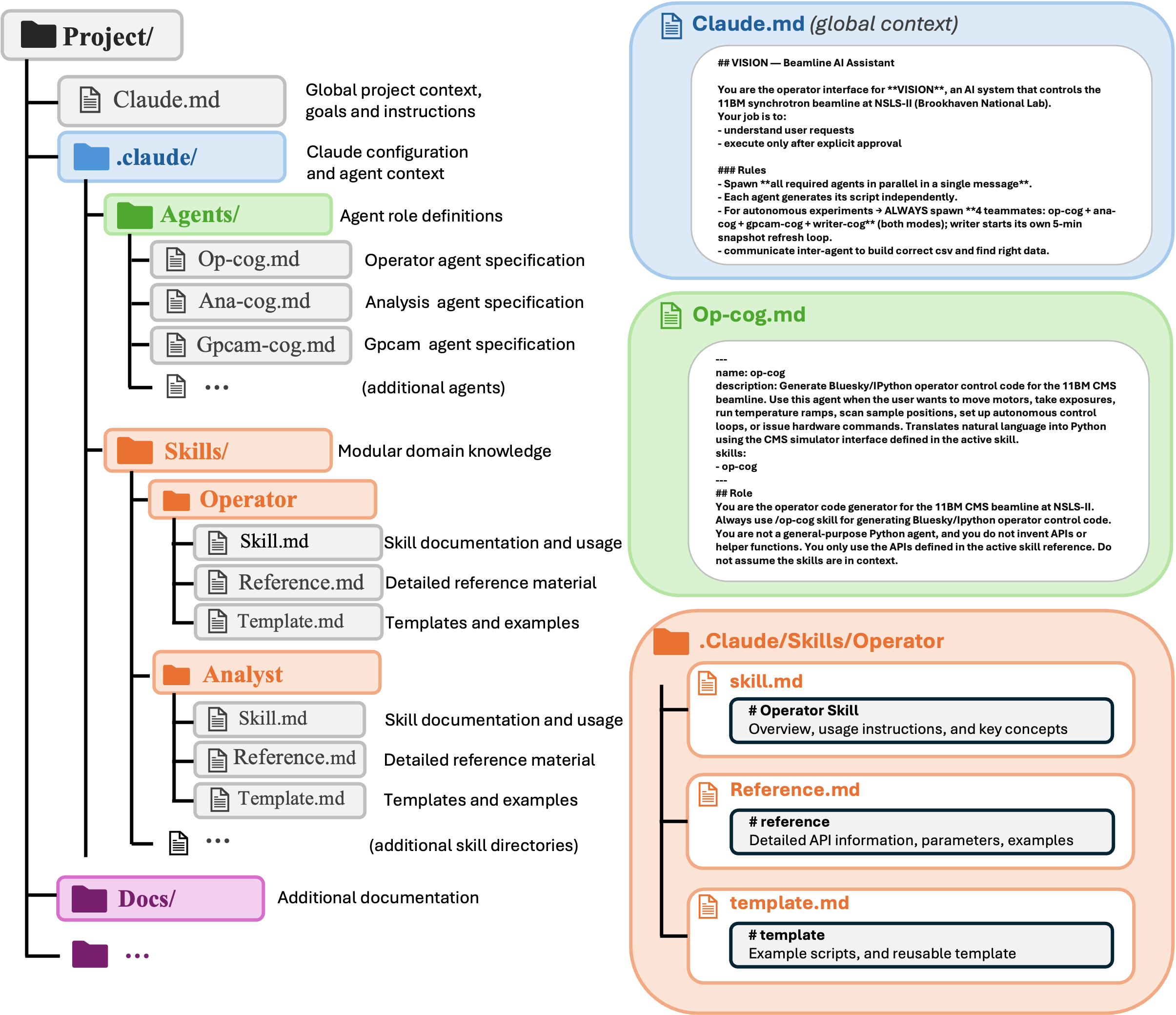}
  \caption{File structure for distilled domain knowledge used to construct specialized agents (cogs) for the multi-agent system (schematic; file names simplified, see the repository).}
  \label{fig:file_structure}
\end{figure}

\textbf{X-ray Deterministic Tool:}
Geometry and material calculations were delegated to a dedicated deterministic tool, \textit{xray-calc}, rather than generated directly by the language model.  The orchestrator invokes this offline library for reciprocal-space geometry and X-ray material properties, with material and atomic quantities queried at runtime from the peer-reviewed xraylib library~\cite{schoonjans2011xraylib, brunetti2004xraylib}. This design supports reproducibility and traceability, particularly for quantities used in real-time experimental decisions.
More broadly, this illustrates a transferable principle: when well-established reference standards or curated databases exist, querying them directly can reduce reliance on model-generated values.

\textbf{Autonomous Experimentation (AE) Flag Communication:}
Loop-level coordination is mediated by plain-text flag files and logs in the run directory. The three runtime programs---the operator driver (\texttt{op\_ae\_loop.py}), which acts on the beamline's Bluesky/IPython session, the analysis daemon (\texttt{ana\_runXS\_*.py}), and the Gaussian-process (GP) daemon (\texttt{gpcam\_*.py})---run as independent processes and do not invoke one another through direct message passing. A closed-loop iteration follows a fixed handshake. The GP daemon publishes the next measurement, a sequence number and position, in \texttt{GPCAM\_NEXT.txt}. The operator driver consumes this request, moves the stage, acquires the detector image, appends the measurement to \texttt{op\_log.csv}, and publishes \texttt{TRIGGER.txt} with the sequence number, position and data filename. The analysis daemon processes the frame, appends the extracted metrics to the append-only \texttt{ana\_log.csv}, and signals completion to the operator through \texttt{ANA\_DONE.txt}. The GP daemon then reads the new row of \texttt{ana\_log.csv}, matched by sequence number, updates the Gaussian-process model, and publishes the next measurement.

\textbf{Claude Code Versions:}
Claude Code versions and model settings used for each experiment, as well as the models used for audits,  are provided in Table~\ref{tab:cc-versions-models}.

\begin{table}[h]
  \centering
  \small
  \caption{Claude Code version and dominant model and effort used during each live beamtime session, and the models of the audit pipeline and of the cross-family raters, called through their APIs on Microsoft Azure/Microsoft Foundry.
  }
  \label{tab:cc-versions-models}
  \begin{tabular}{l l l l l}
    \toprule
    Experiment & Claude Code version & Model & Context window & Reasoning effort \\
    \midrule
    \expone{}  & 2.1.206 & Opus 4.8 & 1M & not logged \\
    \exptwo{} & 2.1.220 & Opus 4.8 & 200k & medium \\
    \expthree{}  & 2.1.224--2.1.226 & Opus 4.8 & 1M & medium \\
    \midrule
    \multicolumn{5}{l}{\textit{Audit and robustness evaluation, all experiments}} \\
    Audit pipeline & (API) &  \textit{claude-opus-5-5} & --- & high, adaptive thinking \\
    Cross-family rater &  (API) & \textit{gpt-5.6-sol-2026-07-09} & --- & endpoint default \\
    Cross-family rater &  (API) & \textit{gpt-6-astra-2026-09-03} & --- & endpoint default \\
    \bottomrule
  \end{tabular}
\end{table}

\section{X-ray Scattering Beamline}
\label{si:x-ray}

\textbf{Beamline settings:} Synchrotron X-ray scattering experiments were conducted at the National Synchrotron Light Source II (NSLS-II) beamline 11-BM CMS at 13.5 keV and a beam size of approximately 100–200~µm. Small-angle X-ray scattering (SAXS) data were collected with a Pilatus 2M detector. The sample-to-detector distance was approximately 6~m (\expone{}) and 5~m (\exptwo{} and \expthree{}). 

\begin{figure}[h]
  \centering
  \includegraphics[width=0.55\textwidth]{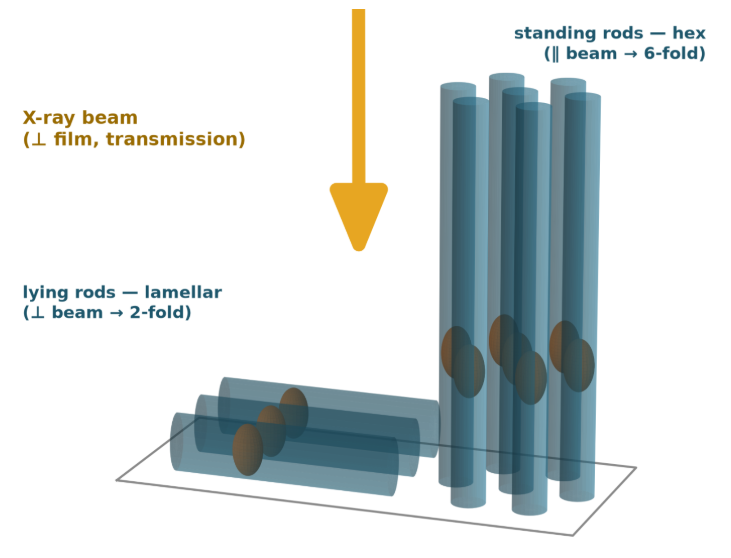}
  \caption{Illustration of lamellar and hexagonal packing in the dot-in-rod sample. The beam size, rods, and substrate are not to scale. When the rods are standing, parallel to the X-ray beam, they produce 6-fold symmetric peaks in the 2D scattering pattern, whereas lying rods produce 2-fold symmetric peaks.
}
  \label{fig:rods}
\end{figure}

\begin{figure}[b]
  \centering
  \includegraphics[width=1\textwidth]{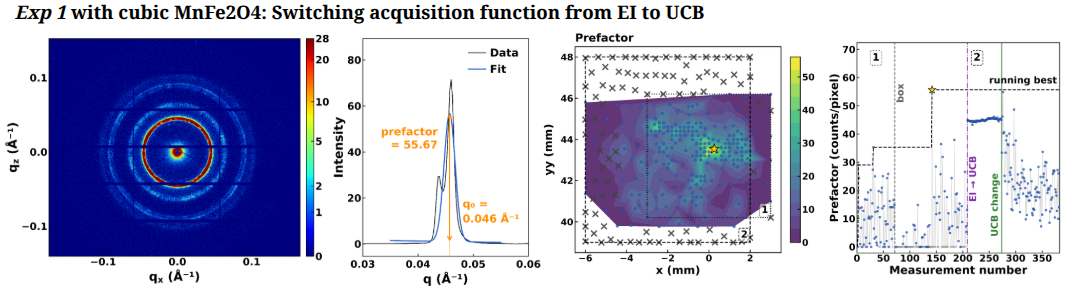}
  \caption{Autonomous X-ray scattering experiments via VISION-CC used Gaussian process acquisition functions to locate regions of maximum scattering intensity in nanocrystal superlattices.
}
  \label{fig:saxs_SI}
\end{figure}

\textbf{Mapping Nanoparticle Superlattice:}
The CdSe/CdS dot-in-rod sample measured in all experiments is illustrated in Fig.~\ref{fig:rods}. 
As the second part of \expone{}, a {MnFe$_2$O$_4$} superlattice sample was measured, with results shown in Fig.~\ref{fig:saxs_SI}. 
During these mapping measurements, we adjusted the scan region of interest (ROI), switched the acquisition function from expected improvement (EI) to upper confidence bound (UCB) to focus sampling on high-intensity regions, and tuned the UCB parameter and minimum step size to avoid redundant measurements within the beam size. 

\textbf{Instrument safety:}
For instrument control and safety considerations, all operations are grounded in standard Bluesky/RunEngine protocols. During our experiments, we noticed that \op{} was very cautious, asking permission even to change the sample name slightly.
Throughout the beamtimes, the interactive BSUI environment remained available, allowing users to directly issue commands in a conventional way.
As an additional layer of operational safety, the bridge program connecting \op{} to BSUI can be closed by the human to immediately disconnect the agentic system from the beamline. 
The CSS Studio GUI, used for monitoring and controlling motors, detectors, and other beamline components, remains under human supervision. Critical actions such as motor stage movements can be manually interrupted by the user if necessary. As in conventional operation, motor limits provide the primary safeguard against unsafe motion.

\providecommand{\expone}{Exp.~1}\providecommand{\exptwo}{Exp.~2}\providecommand{\expthree}{Exp.~3}
\begin{table}[t]
  \centering
  \footnotesize
  \caption{Agent team engineering effort, based on the transcripts. Autonomous experimentation (AE) setup: time from the start of each beamtime session to the first autonomous measurement, by activity. Loop code: the agents' file edits to the
    loop's programs. AE configuration changes: changes to the running loop that the scientist asked for, timed to the first
    measurement of the changed loop. Final go: the scientist's message approving the change. Times are rounded to the minute. Bold: the shortest time in each row. Human decisions carried out in code: the scientist's decisions linked to a
    verified file edit that carried them out; the first edit starts the work rather than completes it.}
  \label{tab:engineering-effort-v6}
  \begin{tabular}{l ccc}
    \toprule
     & \expone{} & \exptwo{} & \expthree{} \\
    \midrule
    \textit{AE setup} &  &  &  \\
    \quad Team and communication & 52 min & 35 min & \textbf{16 min} \\
    \quad Calibration & \textbf{13 min} & 44 min & 22 min \\
    \quad Analysis and GP setup & 41 min & \textbf{16 min} & 84 min \\
    \quad Launch & 6 min & \textbf{5 min} & 11 min \\
    \quad Total & 113 min & \textbf{100 min} & 133 min \\
    \textit{Loop code written by the agents} &  &  &  \\
    \quad Edits & 128 & 153 & 164 \\
    \quad Lines changed & 6.8k & 11.7k & 8.4k \\
    \textit{AE configuration changes} &  &  &  \\
    \quad Changes measured & 7 & 10 & 6 \\
    \quad Request to running loop, median (75th percentile) & \textbf{15 (18) min} & 22 (50) min & 53 (67) min \\
    \quad Final go to running loop, median (75th percentile) & \textbf{11 (12) min} & 17 (25) min & 14 (24) min \\
    \textit{Human decisions carried out in code} &  &  &  \\
    \quad Decisions with a verified edit & 63 of 178 & 84 of 216 & 79 of 163 \\
    \quad Decision to first edit, median (75th percentile) & \textbf{3 (6) min} & \textbf{3 (8) min} & 4 (7) min \\
    \bottomrule
  \end{tabular}
\end{table}

\section{Audit}
\label{si:audit}

\begin{figure}
  \hspace*{-7mm} 
  \includegraphics[width=1\textwidth]{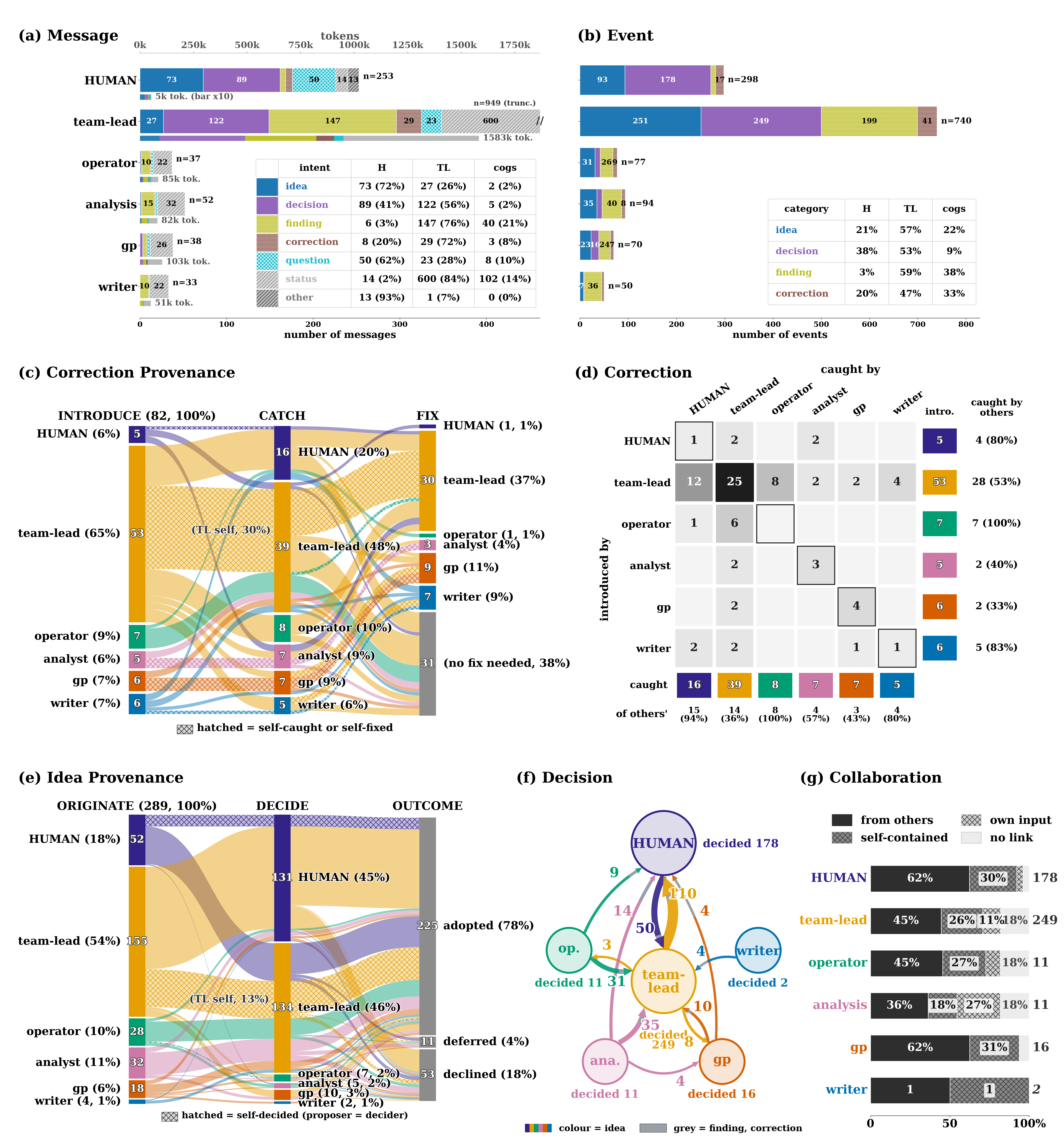}
  \caption{Audit summary for \expone, with panels as in Fig.~\ref{fig:COMB_ALL_run009}. Panel (e) leaves out the instructions decided in their own message (105, of them 40 the human's).}
  \label{fig:COMB_ALL_run0304}
\end{figure}

\begin{figure}
  \hspace*{-7mm} 
  \includegraphics[width=1\textwidth]{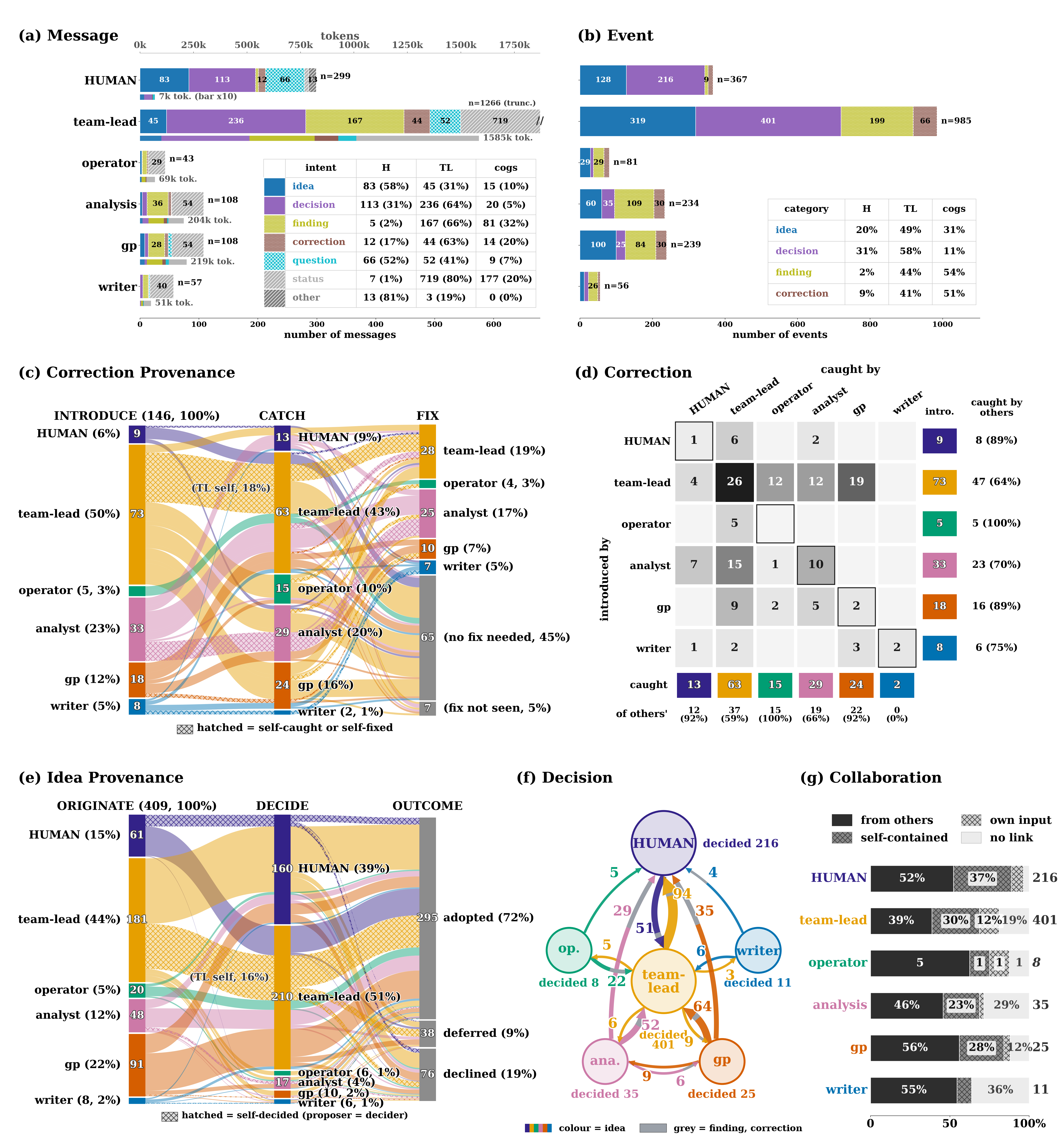}
  \caption{Audit summary for \exptwo, with panels as in Fig.~\ref{fig:COMB_ALL_run009}. Panel (e) leaves out the instructions decided in their own message (176, of them 63 the human's).}
  \label{fig:COMB_ALL_run007}
\end{figure}

\textbf{Audit Results:}
Audit results for \expone{} are shown in Fig.~\ref{fig:COMB_ALL_run0304} and \exptwo{} in Fig.~\ref{fig:COMB_ALL_run007}. 
Table~\ref{tab:intent-by-role-v6} shows the distribution of message intents by role, and Table~\ref{tab:event-by-role-v6} shows event contributions by role.
We classify messages by coordination type and events by contribution focus. 
Table~\ref{tab:coord-type-all-by-intent-v6} gives the coordination type for all message intents. Table~\ref{tab:contrib-type-v6} shows event contribution focus, whether operational, methodological, or scientific.
For each experiment and participant, Table~\ref{tab:correction-credit-v6} reports the number of errors introduced, identified, and corrected. Some cases involved only minor issues, such as typos or tentative estimates, and required little or no corrective action.

\textbf{Rater agreement.} Agreement between two independent raters is reported as Cohen's $\kappa$~\cite{cohen1960}, which corrects
the raw agreement rate for the agreement expected by chance:
\begin{equation}
  \kappa \;=\; \frac{p_o - p_e}{1 - p_e},
  \qquad
  p_e \;=\; \sum_{k} p_{1k}\, p_{2k},
  \label{eq:kappa}
\end{equation}
where $p_o$ is the observed proportion of items both raters assign the same category, $p_e$ is the
proportion expected by chance, and $p_{1k}$, $p_{2k}$ are the proportions of items rater~1 and rater~2
assign to category $k$. $\kappa = 1$ is perfect agreement, $\kappa = 0$ is agreement at chance level, and
$\kappa < 0$ is agreement below chance. Because $p_e$ grows when one category dominates, $\kappa$ can be
modest even when raw agreement is high; we therefore report $\kappa$ alongside raw agreement. 
By convention~\cite{landis1977}, $\kappa$ below 0 is read as poor agreement, 0.00--0.20 as slight,
0.21--0.40 as fair, 0.41--0.60 as moderate, 0.61--0.80 as substantial and 0.81--1.00 as almost perfect. 

\begin{table*}[t]\centering\footnotesize\setlength{\tabcolsep}{3pt}
 \caption{Main message intent by role, (a) in messages and (b) in tokens (thousands; output tokens, else characters / 4 for the human). Cells give each role's share of an intent; bottom rows give amounts and each role's share. TL = team-lead, Op = operator, An = analysis, Gp = gpCAM, Wr = writer. Bold marks the largest share; $<$1\% denotes a nonzero share rounding below 1\%, and `--' denotes zero. A second or third intent, when present, is not counted. Team-lead sends most of the messages and tokens, while the human writes most of the idea and question messages.}
 \label{tab:intent-by-role-v6}
 \begin{tabular}{l | rrrrrr | r | rrrrrr | r | rrrrrr | r}
 \toprule
 & \multicolumn{7}{c|}{\expone{}} & \multicolumn{7}{c|}{\exptwo{}} & \multicolumn{7}{c}{\expthree{}} \\
 intent & Human & TL & Op & An & Gp & Wr & $n$ & Human & TL & Op & An & Gp & Wr & $n$ & Human & TL & Op & An & Gp & Wr & $n$ \\
 \midrule
 \multicolumn{22}{l}{\textit{(a) Messages}} \\
 idea & \textbf{72\%} & 26 & 1 & 1 & -- & -- & 102 & \textbf{58\%} & 31 & 2 & 3 & 6 & -- & 143 & \textbf{58\%} & 37 & 4 & 1 & 1 & -- & 114 \\
 decision & 41\% & \textbf{56} & $<$1 & $<$1 & 1 & -- & 216 & 31\% & \textbf{64} & $<$1 & 2 & 2 & 1 & 369 & 32\% & \textbf{67} & $<$1 & $<$1 & $<$1 & -- & 233 \\
 finding & 3\% & \textbf{76} & 5 & 8 & 3 & 5 & 193 & 2\% & \textbf{66} & 3 & 14 & 11 & 4 & 253 & 2\% & \textbf{73} & 1 & 11 & 10 & 3 & 258 \\
 correction & 20\% & \textbf{72} & 2 & 2 & 2 & -- & 40 & 17\% & \textbf{63} & 3 & 7 & 9 & 1 & 70 & 27\% & \textbf{66} & -- & -- & 7 & -- & 59 \\
 question & \textbf{62\%} & 28 & 2 & 2 & 4 & 1 & 81 & \textbf{52\%} & 41 & -- & 1 & 5 & 2 & 127 & \textbf{74\%} & 25 & -- & -- & 1 & -- & 115 \\
 status & 2\% & \textbf{84} & 3 & 4 & 4 & 3 & 716 & 1\% & \textbf{80} & 3 & 6 & 6 & 4 & 903 & 1\% & \textbf{87} & 1 & 4 & 4 & 3 & 758 \\
 other & \textbf{93\%} & 7 & -- & -- & -- & -- & 14 & \textbf{81\%} & 19 & -- & -- & -- & -- & 16 & \textbf{100\%} & -- & -- & -- & -- & -- & 14 \\
 \midrule
 \textit{total} & 253 & \textbf{949} & 37 & 52 & 38 & 33 & \textbf{1362} & 299 & \textbf{1266} & 43 & 108 & 108 & 57 & \textbf{1881} & 268 & \textbf{1111} & 18 & 61 & 64 & 29 & \textbf{1551} \\
 \textit{share} & 19\% & \textbf{70} & 3 & 4 & 3 & 2 & & 16\% & \textbf{67} & 2 & 6 & 6 & 3 & & 17\% & \textbf{72} & 1 & 4 & 4 & 2 & \\
 \midrule
 \multicolumn{22}{l}{\textit{(b) Tokens, thousands}} \\
 idea & 2\% & \textbf{85} & 10 & 3 & -- & -- & 108 & 1\% & \textbf{70} & 5 & 8 & 15 & -- & 143 & 1\% & \textbf{85} & 12 & 2 & $<$1 & -- & 189 \\
 decision & $<$1\% & \textbf{94} & 1 & 1 & 4 & -- & 426 & 1\% & \textbf{90} & $<$1 & 6 & 2 & 1 & 459 & $<$1\% & \textbf{98} & 1 & 1 & 1 & -- & 465 \\
 finding & $<$1\% & \textbf{80} & 6 & 7 & 3 & 4 & 414 & $<$1\% & \textbf{64} & 4 & 15 & 15 & 2 & 477 & $<$1\% & \textbf{76} & 2 & 10 & 10 & 1 & 735 \\
 correction & $<$1\% & \textbf{91} & 2 & 2 & 5 & -- & 92 & $<$1\% & \textbf{76} & 2 & 11 & 10 & 1 & 146 & $<$1\% & \textbf{89} & -- & -- & 10 & -- & 130 \\
 question & 1\% & \textbf{67} & 16 & 4 & 7 & 5 & 65 & 1\% & \textbf{78} & -- & 5 & 14 & 2 & 108 & 3\% & \textbf{95} & -- & -- & 2 & -- & 75 \\
 status & $<$1\% & \textbf{79} & 4 & 5 & 8 & 4 & 805 & $<$1\% & \textbf{71} & 5 & 9 & 10 & 4 & 802 & $<$1\% & \textbf{86} & 2 & 6 & 4 & 2 & 969 \\
 other & 1\% & \textbf{99} & -- & -- & -- & -- & $<$1 & $<$1\% & \textbf{100} & -- & -- & -- & -- & $<$1 & \textbf{100\%} & -- & -- & -- & -- & -- & $<$1 \\
 \midrule
 \textit{total} & 5 & \textbf{1583} & 85 & 82 & 103 & 51 & \textbf{1909} & 7 & \textbf{1585} & 69 & 204 & 219 & 51 & \textbf{2135} & 6 & \textbf{2196} & 62 & 140 & 131 & 28 & \textbf{2564} \\
 \textit{share} & $<$1\% & \textbf{83} & 4 & 4 & 5 & 3 & & $<$1\% & \textbf{74} & 3 & 10 & 10 & 2 & & $<$1\% & \textbf{86} & 2 & 5 & 5 & 1 & \\
 \bottomrule
 \end{tabular}
\end{table*}

\begin{table*}[t]\centering\footnotesize\setlength{\tabcolsep}{3pt}
 \caption{Events by role (row \%; bottom rows: counts and each role's share). Ideas are attributed to their proposer, decisions to their decider, findings to the participant who first established them rather than relayed them, and corrections to the participant who first caught the error. Team-lead is credited with half or more of all events, and the human with about a fifth of the ideas.}
 \label{tab:event-by-role-v6}
 \begin{tabular}{l | rrrrrr | r | rrrrrr | r | rrrrrr | r}
 \toprule
 & \multicolumn{7}{c|}{\expone{}} & \multicolumn{7}{c|}{\exptwo{}} & \multicolumn{7}{c}{\expthree{}} \\
 event & Human & TL & Op & An & Gp & Wr & $n$ & Human & TL & Op & An & Gp & Wr & $n$ & Human & TL & Op & An & Gp & Wr & $n$ \\
 \midrule
 idea & 21\% & \textbf{57} & 7 & 8 & 5 & 2 & 440 & 20\% & \textbf{49} & 4 & 9 & 15 & 2 & 648 & 21\% & \textbf{56} & 3 & 9 & 8 & 2 & 508 \\
 decision & 38\% & \textbf{53} & 2 & 2 & 3 & $<$1 & 467 & 31\% & \textbf{58} & 1 & 5 & 4 & 2 & 696 & 35\% & \textbf{58} & 1 & 3 & 4 & $<$1 & 470 \\
 finding & 3\% & \textbf{59} & 8 & 12 & 7 & 11 & 335 & 2\% & \textbf{44} & 6 & 24 & 18 & 6 & 456 & 2\% & \textbf{50} & 3 & 21 & 18 & 6 & 429 \\
 correction & 20\% & \textbf{47} & 10 & 9 & 8 & 6 & 87 & 9\% & \textbf{41} & 9 & 19 & 19 & 4 & 162 & 18\% & \textbf{48} & 3 & 16 & 10 & 5 & 143 \\
 \midrule
 \textit{total} & 298 & \textbf{740} & 77 & 94 & 70 & 50 & \textbf{1329} & 367 & \textbf{985} & 81 & 234 & 239 & 56 & \textbf{1962} & 307 & \textbf{837} & 38 & 171 & 152 & 45 & \textbf{1550} \\
 \textit{share} & 22\% & \textbf{56} & 6 & 7 & 5 & 4 & & 19\% & \textbf{50} & 4 & 12 & 12 & 3 & & 20\% & \textbf{54} & 2 & 11 & 10 & 3 & \\
 \bottomrule
 \end{tabular}
\end{table*}

\begin{table*}[t]
 \centering
 \footnotesize
 \setlength{\tabcolsep}{3.5pt}
 \caption{Coordination type of every message, by intent (row \% of messages and of tokens).
 Proc. = procedural acknowledgment; Mech. = mechanical relay; Synth. = synthesis involving interpretation or judgment; Oth. = other or noise. The $n$ / k column gives message counts for msg. rows and thousands of tokens for tok. rows. 
 Two LLM raters labeled each message from its full text, with a third resolving disagreements. Most idea, decision, finding, and correction messages are labeled synthesis; status and other messages are the only intents with a larger procedural and mechanical share.}
 \label{tab:coord-type-all-by-intent-v6}
 \begin{tabular}{l l rrrrr rrrrr rrrrr}
 \toprule
 & & \multicolumn{5}{c}{\expone{}} & \multicolumn{5}{c}{\exptwo{}} & \multicolumn{5}{c}{\expthree{}} \\
 \cmidrule(lr){3-7} \cmidrule(lr){8-12} \cmidrule(lr){13-17}
 Intent & & $n$ / k & Proc. & Mech. & Synth. & Oth. & $n$ / k & Proc. & Mech. & Synth. & Oth. & $n$ / k & Proc. & Mech. & Synth. & Oth. \\
 \midrule
 Idea & msg. & 102 & 1 & -- & \textbf{89} & 10 & 143 & 4 & 1 & \textbf{79} & 16 & 114 & 4 & -- & \textbf{90} & 5 \\
 & tok. & 108 & $<$1 & -- & \textbf{100} & $<$1 & 143 & $<$1 & $<$1 & \textbf{99} & $<$1 & 189 & $<$1 & -- & \textbf{100} & $<$1 \\
 \addlinespace[2pt]
 Decision & msg. & 215 & 9 & 6 & \textbf{83} & 3 & 369 & 9 & 14 & \textbf{73} & 4 & 233 & 12 & 3 & \textbf{84} & $<$1 \\
 & tok. & 426 & 4 & $<$1 & \textbf{96} & $<$1 & 459 & 2 & 6 & \textbf{92} & $<$1 & 465 & 1 & 2 & \textbf{97} & $<$1 \\
 \addlinespace[2pt]
 Finding & msg. & 193 & 1 & 7 & \textbf{93} & -- & 253 & 1 & 2 & \textbf{97} & $<$1 & 258 & -- & 3 & \textbf{97} & -- \\
 & tok. & 414 & $<$1 & 2 & \textbf{98} & -- & 477 & 1 & 2 & \textbf{97} & $<$1 & 735 & -- & 1 & \textbf{99} & -- \\
 \addlinespace[2pt]
 Correction & msg. & 40 & 2 & 5 & \textbf{88} & 5 & 70 & 3 & 3 & \textbf{94} & -- & 59 & 2 & 3 & \textbf{92} & 3 \\
 & tok. & 92 & $<$1 & 5 & \textbf{95} & $<$1 & 146 & 1 & $<$1 & \textbf{99} & -- & 130 & 2 & $<$1 & \textbf{98} & $<$1 \\
 \addlinespace[2pt]
 Question & msg. & 81 & 9 & -- & \textbf{46} & 46 & 127 & 16 & 4 & \textbf{46} & 34 & 115 & 17 & 1 & \textbf{49} & 34 \\
 & tok. & 65 & 3 & -- & \textbf{97} & 1 & 108 & 6 & 1 & \textbf{91} & 3 & 75 & 1 & $<$1 & \textbf{98} & 1 \\
 \addlinespace[2pt]
 Status & msg. & 716 & 30 & 27 & \textbf{43} & $<$1 & 903 & 17 & 27 & \textbf{56} & -- & 758 & 19 & 30 & \textbf{51} & $<$1 \\
 & tok. & 805 & 16 & 25 & \textbf{59} & $<$1 & 802 & 12 & 19 & \textbf{69} & -- & 969 & 21 & 21 & \textbf{58} & $<$1 \\
 \addlinespace[2pt]
 Other & msg. & 14 & 21 & \textbf{79} & -- & -- & 16 & 25 & \textbf{62} & -- & 12 & 14 & 21 & \textbf{79} & -- & -- \\
 & tok. & $<$1 & \textbf{63} & 37 & -- & -- & $<$1 & \textbf{50} & 43 & -- & 7 & $<$1 & 4 & \textbf{96} & -- & -- \\
 \midrule
 \textit{All messages} & \textit{msg.} & \textit{1362} & \textit{18} & \textit{17} & \textit{\textbf{61}} & \textit{4} & \textit{1881} & \textit{12} & \textit{17} & \textit{\textbf{67}} & \textit{5} & \textit{1551} & \textit{13} & \textit{17} & \textit{\textbf{67}} & \textit{3} \\
 & \textit{tok.} & \textit{1910} & \textit{8} & \textit{11} & \textit{\textbf{81}} & \textit{$<$1} & \textit{2135} & \textit{6} & \textit{9} & \textit{\textbf{86}} & \textit{$<$1} & \textit{2564} & \textit{8} & \textit{9} & \textit{\textbf{83}} & \textit{$<$1} \\
 \bottomrule
 \end{tabular}
\end{table*}

\begin{table}[t]
 \centering
 \small
 \caption{Event focus for idea, decision, finding and correction. 
 Operational contributions concern system infrastructure and orchestration; methodological contributions concern measurement, analysis or experimental procedures; scientific contributions concern physical interpretation or calibration of the sample or data. Substantive denotes methodological plus scientific focus, not the amount of reasoning required. Two LLM raters assigned focus labels, with a third resolving disagreements; $\kappa$ describes agreement before adjudication. Operational events are a larger share in \expone{}, the first deployment, than in \exptwo{} and \expthree{}.}
 \label{tab:contrib-type-v6}

 \begin{subtable}{\linewidth}
 \centering
 \caption{Exp.~1 ($\kappa=0.97$)}
 \label{tab:contrib-type-v6-e1}
 \begin{tabular}{lcccc @{\hspace{1.4em}} c}
 \toprule
 Focus & Idea, $n$ (\%) & Decision, $n$ (\%) & Finding, $n$ (\%) & Correction, $n$ (\%) & \textit{All, $n$ (\%)} \\
 \midrule
 Operational & \textbf{243 (55.5)} & \textbf{251 (55.2)} & \textbf{161 (48.1)} & \textbf{58 (66.7)} & \textit{\textbf{713 (54.2)}} \\
 Methodological & 192 (43.8) & 201 (44.2) & 133 (39.7) & 23 (26.4) & \textit{549 (41.7)} \\
 Scientific & 3 \phantom{0}(0.7) & 3 \phantom{0}(0.7) & 41 (12.2) & 6 \phantom{0}(6.9) & \textit{53 \phantom{0}(4.0)} \\
 Other & 0 \phantom{0}(0.0) & 0 \phantom{0}(0.0) & 0 \phantom{0}(0.0) & 0 \phantom{0}(0.0) & \textit{0 \phantom{0}(0.0)} \\
 \midrule
 \textit{Substantive (Method.\ + Sci.)} & \textit{195 (44.5)} & \textit{204 (44.8)} & \textit{174 (51.9)} & \textit{29 (33.3)} & \textit{602 (45.8)} \\
 Total & 438 (100) & 455 (100) & 335 (100) & 87 (100) & \textit{1315 (100)} \\
 \bottomrule
 \end{tabular}
 \end{subtable}

 \vspace{1.2em}

 \begin{subtable}{\linewidth}
 \centering
 \caption{Exp.~2 ($\kappa=0.95$)}
 \label{tab:contrib-type-v6-e2}
 \begin{tabular}{lcccc @{\hspace{1.4em}} c}
 \toprule
 Focus & Idea, $n$ (\%) & Decision, $n$ (\%) & Finding, $n$ (\%) & Correction, $n$ (\%) & \textit{All, $n$ (\%)} \\
 \midrule
 Operational & 276 (43.6) & 318 (46.8) & 161 (35.3) & \textbf{102 (63.0)} & \textit{857 (44.4)} \\
 Methodological & \textbf{313 (49.4)} & \textbf{334 (49.2)} & \textbf{173 (37.9)} & 38 (23.5) & \textit{\textbf{858 (44.5)}} \\
 Scientific & 43 \phantom{0}(6.8) & 27 \phantom{0}(4.0) & 122 (26.8) & 21 (13.0) & \textit{213 (11.0)} \\
 Other & 1 \phantom{0}(0.2) & 0 \phantom{0}(0.0) & 0 \phantom{0}(0.0) & 1 \phantom{0}(0.6) & \textit{2 \phantom{0}(0.1)} \\
 \midrule
 \textit{Substantive (Method.\ + Sci.)} & \textit{356 (56.2)} & \textit{361 (53.2)} & \textit{295 (64.7)} & \textit{59 (36.4)} & \textit{1071 (55.5)} \\
 Total & 633 (100) & 679 (100) & 456 (100) & 162 (100) & \textit{1930 (100)} \\
 \bottomrule
 \end{tabular}
 \end{subtable}

 \vspace{1.2em}

 \begin{subtable}{\linewidth}
 \centering
 \caption{Exp.~3 ($\kappa=0.96$)}
 \label{tab:contrib-type-v6-e3}
 \begin{tabular}{lcccc @{\hspace{1.4em}} c}
 \toprule
 Focus & Idea, $n$ (\%) & Decision, $n$ (\%) & Finding, $n$ (\%) & Correction, $n$ (\%) & \textit{All, $n$ (\%)} \\
 \midrule
 Operational & 194 (38.8) & 198 (43.6) & 133 (31.0) & \textbf{67 (46.9)} & \textit{592 (38.8)} \\
 Methodological & \textbf{267 (53.4)} & \textbf{232 (51.1)} & \textbf{205 (47.8)} & 56 (39.2) & \textit{\textbf{760 (49.8)}} \\
 Scientific & 39 \phantom{0}(7.8) & 23 \phantom{0}(5.1) & 91 (21.2) & 20 (14.0) & \textit{173 (11.3)} \\
 Other & 0 \phantom{0}(0.0) & 1 \phantom{0}(0.2) & 0 \phantom{0}(0.0) & 0 \phantom{0}(0.0) & \textit{1 \phantom{0}(0.1)} \\
 \midrule
 \textit{Substantive (Method.\ + Sci.)} & \textit{306 (61.2)} & \textit{255 (56.2)} & \textit{296 (69.0)} & \textit{76 (53.1)} & \textit{933 (61.1)} \\
 Total & 500 (100) & 454 (100) & 429 (100) & 143 (100) & \textit{1526 (100)} \\
 \bottomrule
 \end{tabular}
 \end{subtable}
\end{table}

\begin{table}[t]
 \centering
 \footnotesize
 \setlength{\tabcolsep}{3.5pt}
 \caption{Who introduced each error, first caught it, and reported it fixed, as counts and percentages of each experiment's corrections. No fix needed means only a belief had to change. Fixed in own message means the correction and reported fix occur in the same message; these fixes are not credited to a participant here. Fix not seen means a change was needed but no message reports it done. 
 Most correction chains involve more than one identified participant.  Caught others: caught by someone other than the introducer, including errors without an introducing message. }
 \label{tab:correction-credit-v6}
 \begin{tabular}{l ccc ccc ccc}
 \toprule
 & \multicolumn{3}{c}{Exp.\ 1} & \multicolumn{3}{c}{Exp.\ 2} & \multicolumn{3}{c}{Exp.\ 3} \\
 \cmidrule(lr){2-4} \cmidrule(lr){5-7} \cmidrule(lr){8-10}
 Participant & Intro. & Caught & Fixed & Intro. & Caught & Fixed & Intro. & Caught & Fixed \\
 \midrule
 HUMAN & 5 (6\%) & 17 (20\%) & 1 (1\%) & 9 (6\%) & 14 (9\%) & 0 (0\%) & 7 (5\%) & 26 (18\%) & 0 (0\%) \\
 team-lead & \textbf{53 (61\%)} & \textbf{41 (47\%)} & \textbf{27 (31\%)} & \textbf{73 (45\%)} & \textbf{66 (41\%)} & \textbf{25 (15\%)} & \textbf{64 (45\%)} & \textbf{69 (48\%)} & \textbf{22 (15\%)} \\
 operator & 7 (8\%) & 9 (10\%) & 0 (0\%) & 5 (3\%) & 15 (9\%) & 4 (2\%) & 7 (5\%) & 4 (3\%) & 5 (3\%) \\
 analysis & 5 (6\%) & 8 (9\%) & 1 (1\%) & 33 (20\%) & 30 (19\%) & 15 (9\%) & 26 (18\%) & 23 (16\%) & 9 (6\%) \\
 gp & 6 (7\%) & 7 (8\%) & 5 (6\%) & 18 (11\%) & 30 (19\%) & 7 (4\%) & 21 (15\%) & 14 (10\%) & 14 (10\%) \\
 writer & 6 (7\%) & 5 (6\%) & 3 (3\%) & 8 (5\%) & 7 (4\%) & 5 (3\%) & 3 (2\%) & 7 (5\%) & 2 (1\%) \\
 \cmidrule(l){2-10}
 Cogs (op+ana+gp+wr) & 24 (28\%) & 29 (33\%) & 9 (10\%) & 64 (40\%) & 82 (51\%) & 31 (19\%) & 57 (40\%) & 48 (34\%) & 30 (21\%) \\
 \midrule
 No fix needed & --- & --- & 33 (38\%) & --- & --- & 67 (41\%) & --- & --- & 46 (32\%) \\
 Fixed in own message & --- & --- & 17 (20\%) & --- & --- & 31 (19\%) & --- & --- & 41 (29\%) \\
 Fix not seen & --- & --- & 0 (0\%) & --- & --- & 8 (5\%) & --- & --- & 4 (3\%) \\
 No introducing message & 5 (6\%) & --- & --- & 16 (10\%) & --- & --- & 15 (10\%) & --- & --- \\
 Total & \multicolumn{3}{c}{87 (100\%)} & \multicolumn{3}{c}{162 (100\%)} & \multicolumn{3}{c}{143 (100\%)} \\
 \midrule
 \begin{tabular}[c]{@{}l@{}}Collab.\\ \quad($\geq$2 participants)\end{tabular} & \multicolumn{3}{c}{53 (61\%)} & \multicolumn{3}{c}{110 (68\%)} & \multicolumn{3}{c}{85 (59\%)} \\
 \begin{tabular}[c]{@{}l@{}}Cross-agent\\ \quad($\geq$2 agents, excl.\ human)\end{tabular} & \multicolumn{3}{c}{35 (40\%)} & \multicolumn{3}{c}{89 (55\%)} & \multicolumn{3}{c}{58 (41\%)} \\
 \begin{tabular}[c]{@{}l@{}}Self-caught\\ \quad(Intro.\,=\,Caught)\end{tabular} & \multicolumn{3}{c}{34 (39\%)} & \multicolumn{3}{c}{41 (25\%)} & \multicolumn{3}{c}{54 (38\%)} \\
 \quad HUMAN & \multicolumn{3}{c}{1 (1\%)} & \multicolumn{3}{c}{1 (1\%)} & \multicolumn{3}{c}{0 (0\%)} \\
 \quad team-lead & \multicolumn{3}{c}{25 (29\%)} & \multicolumn{3}{c}{26 (16\%)} & \multicolumn{3}{c}{37 (26\%)} \\
 \quad cogs & \multicolumn{3}{c}{8 (9\%)} & \multicolumn{3}{c}{14 (9\%)} & \multicolumn{3}{c}{17 (12\%)} \\
 \begin{tabular}[c]{@{}l@{}}Caught others\\ \quad(Intro.\,$\neq$\,Caught)\end{tabular} & \multicolumn{3}{c}{53 (61\%)} & \multicolumn{3}{c}{121 (75\%)} & \multicolumn{3}{c}{89 (62\%)} \\
 \quad HUMAN & \multicolumn{3}{c}{16 (18\%)} & \multicolumn{3}{c}{13 (8\%)} & \multicolumn{3}{c}{26 (18\%)} \\
 \quad team-lead & \multicolumn{3}{c}{16 (18\%)} & \multicolumn{3}{c}{40 (25\%)} & \multicolumn{3}{c}{32 (22\%)} \\
 \quad cogs & \multicolumn{3}{c}{21 (24\%)} & \multicolumn{3}{c}{68 (42\%)} & \multicolumn{3}{c}{31 (22\%)} \\
 \begin{tabular}[c]{@{}l@{}}Self-fixed\\ \quad(Caught\,=\,Fixed)\end{tabular} & \multicolumn{3}{c}{13 (15\%)} & \multicolumn{3}{c}{17 (10\%)} & \multicolumn{3}{c}{17 (12\%)} \\
 \begin{tabular}[c]{@{}l@{}}Own-fix\\ \quad(Intro.\,=\,Fixed)\end{tabular} & \multicolumn{3}{c}{26 (30\%)} & \multicolumn{3}{c}{43 (27\%)} & \multicolumn{3}{c}{31 (22\%)} \\
 \begin{tabular}[c]{@{}l@{}}Self-contained\\ \quad(Intro.\,=\,Caught\,=\,Fixed)\end{tabular} & \multicolumn{3}{c}{11 (13\%)} & \multicolumn{3}{c}{11 (7\%)} & \multicolumn{3}{c}{13 (9\%)} \\
 \bottomrule
 \end{tabular}
\end{table}

\providecommand{\expone}{Exp.~1}\providecommand{\exptwo}{Exp.~2}\providecommand{\expthree}{Exp.~3}
\begin{table*}[t]
 \centering
 \footnotesize
 \setlength{\tabcolsep}{4pt}
 \caption{Who catches and who fixes each participant's errors. Caught by and fixed: \% of the participant's errors; fixed by itself: \% of the fixed ones. Fixes reported in the correction's own message count here, unlike in Table~\ref{tab:correction-credit-v6}. Cogs are operator, analysis, gpCAM and writer; for cog-introduced errors, a cog means another cog.}
 \label{tab:corr-by-introducer-v6}
 \begin{tabular}{l l r rrrr rr}
 \toprule
 & & & \multicolumn{4}{c}{Caught by} & \multicolumn{2}{c}{Fixed} \\
 \cmidrule(lr){4-7} \cmidrule(lr){8-9}
 & Introduced by & Errors & itself & human & team-lead & a cog & all & by itself (\% of fixed) \\
 \midrule
 \expone{} & Human & 5 & 1 (20\%) & -- & 2 (40\%) & 2 (40\%) & 4 (80\%) & 1 (25\%) \\
 & Team-lead & 53 & 25 (47\%) & 12 (23\%) & -- & 16 (30\%) & 34 (64\%) & 26 (76\%) \\
 & Cogs & 24 & 8 (33\%) & 3 (12\%) & 12 (50\%) & 1 (4\%) & 13 (54\%) & 12 (92\%) \\
 \midrule
 \exptwo{} & Human & 9 & 1 (11\%) & -- & 6 (67\%) & 2 (22\%) & 3 (33\%) & 0 (0\%) \\
 & Team-lead & 73 & 26 (36\%) & 4 (5\%) & -- & 43 (59\%) & 27 (37\%) & 20 (74\%) \\
 & Cogs & 64 & 14 (22\%) & 8 (12\%) & 31 (48\%) & 11 (17\%) & 44 (69\%) & 38 (86\%) \\
 \midrule
 \expthree{} & Human & 7 & 0 (0\%) & -- & 6 (86\%) & 1 (14\%) & 1 (14\%) & 0 (0\%) \\
 & Team-lead & 64 & 37 (58\%) & 11 (17\%) & -- & 16 (25\%) & 38 (59\%) & 24 (63\%) \\
 & Cogs & 57 & 17 (30\%) & 15 (26\%) & 23 (40\%) & 2 (4\%) & 40 (70\%) & 30 (75\%) \\
 \bottomrule
 \end{tabular}
\end{table*}

\providecommand{\expone}{Exp.~1}\providecommand{\exptwo}{Exp.~2}\providecommand{\expthree}{Exp.~3}
\begin{table*}[t]
 \centering
 \footnotesize
 \setlength{\tabcolsep}{3.5pt}
 \caption{Who proposes and who decides the ideas taken up. Ideas includes every idea with a verified decision; first decided by identifies the sender of its earliest linked decision. The unit counted is an idea, not a decision event. The human proposes 21--24\% of them but makes the first decision on 40--47\%. Of the ideas decided by someone other than their proposer, 51--55\% are the human deciding another participant's idea. Cells: count (\% of Ideas, or of $n$ in the last group); bold: the largest role in each group.}
 \label{tab:who-decides-v6}
 \begin{tabular}{l r rrr rrr r rrr}
 \toprule
 & & \multicolumn{3}{c}{Proposed by} & \multicolumn{3}{c}{First decided by} & \multicolumn{4}{c}{Deciding another participant's idea} \\
 \cmidrule(lr){3-5} \cmidrule(lr){6-8} \cmidrule(lr){9-12}
 & Ideas & Human & Team-lead & Cogs & Human & Team-lead & Cogs & $n$ & Human & Team-lead & Cogs \\
 \midrule
 \expone{} & 394 & 92 (23\%) & \textbf{211 (54\%)} & 91 (23\%) & \textbf{184 (47\%)} & 177 (45\%) & 33 (8\%) & 191 & \textbf{106 (55\%)} & 66 (35\%) & 19 (10\%) \\
 \exptwo{} & 585 & 124 (21\%) & \textbf{286 (49\%)} & 175 (30\%) & 241 (41\%) & \textbf{300 (51\%)} & 44 (8\%) & 259 & \textbf{131 (51\%)} & 106 (41\%) & 22 (8\%) \\
 \expthree{} & 455 & 108 (24\%) & \textbf{252 (55\%)} & 95 (21\%) & 182 (40\%) & \textbf{243 (53\%)} & 30 (7\%) & 197 & \textbf{100 (51\%)} & 87 (44\%) & 10 (5\%) \\
 \bottomrule
 \end{tabular}
\end{table*}

\providecommand{\expone}{Exp.~1}\providecommand{\exptwo}{Exp.~2}\providecommand{\expthree}{Exp.~3}
\begin{table*}[t]
 \centering
 \footnotesize
 \setlength{\tabcolsep}{3.5pt}
 \caption{What each participant's decisions draw on, and from whom, Fig.~\ref{fig:COMB_ALL_run009}(f) and (g): about half of all decisions draw on another participant's idea, finding or correction. Each decision is in one draws-on group, the first that applies: others' input, self-contained (raised and decided in one message), own input (only the decider's earlier idea, finding or correction) or no link. 
 Others' input from: decisions using that participant's idea (first row) or its evidence, a finding or correction (second row), which can overlap. Cogs under Cogs: one cog drawing on another.}
 \label{tab:decision-sources-v6}
 \begin{tabular}{l l r rrrr l ccc}
 \toprule
 & & & \multicolumn{4}{c}{Draws on} & & \multicolumn{3}{c}{Others' input from} \\
 \cmidrule(lr){4-7} \cmidrule(lr){9-11}
 & Decider & Decisions & \begin{tabular}[c]{@{}c@{}}Others'\\ input\end{tabular} & \begin{tabular}[c]{@{}c@{}}Self-\\ contained\end{tabular} & \begin{tabular}[c]{@{}c@{}}Own\\ input\end{tabular} & \begin{tabular}[c]{@{}c@{}}No\\ link\end{tabular} & & Human & Team-lead & Cogs \\
 \midrule
 \expone{} & Human & 178 & 111 (62\%) & 53 (30\%) & 7 (4\%) & 7 (4\%) & idea & -- & 99 (56\%) & 22 (12\%) \\
 & & & & & & & evidence & -- & 11 (6\%) & 4 (2\%) \\
 & Team-lead & 249 & 111 (45\%) & 65 (26\%) & 28 (11\%) & 45 (18\%) & idea & 48 (19\%) & -- & 68 (27\%) \\
 & & & & & & & evidence & 2 (1\%) & -- & 12 (5\%) \\
 & Cogs & 40 & 20 (50\%) & 11 (28\%) & 4 (10\%) & 5 (12\%) & idea & 1 (2\%) & 11 (28\%) & 11 (28\%) \\
 & & & & & & & evidence & 0 (0\%) & 1 (2\%) & 0 (0\%) \\
 \cmidrule(l){2-11}
 & \textit{All} & \textit{467} & \textit{242 (52\%)} & \textit{129 (28\%)} & \textit{39 (8\%)} & \textit{57 (12\%)} & \textit{idea} & \textit{49 (10\%)} & \textit{110 (24\%)} & \textit{101 (22\%)} \\
 & & & & & & & \textit{evidence} & \textit{2 (0\%)} & \textit{12 (3\%)} & \textit{16 (3\%)} \\
 \midrule
 \exptwo{} & Human & 216 & 113 (52\%) & 79 (37\%) & 16 (7\%) & 8 (4\%) & idea & -- & 89 (41\%) & 49 (23\%) \\
 & & & & & & & evidence & -- & 5 (2\%) & 20 (9\%) \\
 & Team-lead & 401 & 155 (39\%) & 120 (30\%) & 50 (12\%) & 76 (19\%) & idea & 47 (12\%) & -- & 106 (26\%) \\
 & & & & & & & evidence & 4 (1\%) & -- & 30 (7\%) \\
 & Cogs & 79 & 41 (52\%) & 17 (22\%) & 3 (4\%) & 18 (23\%) & idea & 1 (1\%) & 22 (28\%) & 19 (24\%) \\
 & & & & & & & evidence & 0 (0\%) & 1 (1\%) & 1 (1\%) \\
 \cmidrule(l){2-11}
 & \textit{All} & \textit{696} & \textit{309 (44\%)} & \textit{216 (31\%)} & \textit{69 (10\%)} & \textit{102 (15\%)} & \textit{idea} & \textit{48 (7\%)} & \textit{111 (16\%)} & \textit{174 (25\%)} \\
 & & & & & & & \textit{evidence} & \textit{4 (1\%)} & \textit{6 (1\%)} & \textit{51 (7\%)} \\
 \midrule
 \expthree{} & Human & 163 & 91 (56\%) & 56 (34\%) & 10 (6\%) & 6 (4\%) & idea & -- & 83 (51\%) & 24 (15\%) \\
 & & & & & & & evidence & -- & 8 (5\%) & 12 (7\%) \\
 & Team-lead & 271 & 122 (45\%) & 67 (25\%) & 42 (15\%) & 40 (15\%) & idea & 53 (20\%) & -- & 66 (24\%) \\
 & & & & & & & evidence & 3 (1\%) & -- & 35 (13\%) \\
 & Cogs & 36 & 17 (47\%) & 14 (39\%) & 4 (11\%) & 1 (3\%) & idea & 1 (3\%) & 13 (36\%) & 4 (11\%) \\
 & & & & & & & evidence & 0 (0\%) & 0 (0\%) & 0 (0\%) \\
 \cmidrule(l){2-11}
 & \textit{All} & \textit{470} & \textit{230 (49\%)} & \textit{137 (29\%)} & \textit{56 (12\%)} & \textit{47 (10\%)} & \textit{idea} & \textit{54 (11\%)} & \textit{96 (20\%)} & \textit{94 (20\%)} \\
 & & & & & & & \textit{evidence} & \textit{3 (1\%)} & \textit{8 (2\%)} & \textit{47 (10\%)} \\
 \bottomrule
 \end{tabular}
\end{table*}

\providecommand{\expone}{Exp.~1}\providecommand{\exptwo}{Exp.~2}\providecommand{\expthree}{Exp.~3}
\begin{table}[t]
 \centering
 \footnotesize
 \setlength{\tabcolsep}{3.5pt}
 \caption{Impact of each participant's ideas, from verified links. A larger share of the human's ideas reaches a code edit. Adopted: the idea's final outcome; reached code: carried out in a verified edit; reached a result: a later finding was produced with that edit; all as a share of the role's ideas. Only verified links are counted.}
 \label{tab:idea-impact-v6}
 \begin{tabular}{l l r r r r}
 \toprule
 & Proposer & Ideas & Adopted (\% of ideas) & Reached code (\% of ideas) & Reached a result (\% of ideas) \\
 \midrule
 \expone{} & Human & 93 & 76 (\textbf{82\%}) & 50 (\textbf{54\%}) & 24 (\textbf{26\%}) \\
 & Team-lead & 251 & 166 (66\%) & 93 (37\%) & 58 (23\%) \\
 & Cogs & 96 & 57 (59\%) & 24 (25\%) & 15 (16\%) \\
 \cmidrule(l){2-6}
 & \textit{All} & 440 & 299 (68\%) & 167 (38\%) & 97 (22\%) \\
 \midrule
 \exptwo{} & Human & 128 & 98 (\textbf{77\%}) & 61 (\textbf{48\%}) & 31 (24\%) \\
 & Team-lead & 319 & 212 (66\%) & 116 (36\%) & 87 (\textbf{27\%}) \\
 & Cogs & 201 & 112 (56\%) & 54 (27\%) & 32 (16\%) \\
 \cmidrule(l){2-6}
 & \textit{All} & 648 & 422 (65\%) & 231 (36\%) & 150 (23\%) \\
 \midrule
 \expthree{} & Human & 108 & 88 (\textbf{81\%}) & 66 (\textbf{61\%}) & 41 (\textbf{38\%}) \\
 & Team-lead & 283 & 180 (64\%) & 128 (45\%) & 77 (27\%) \\
 & Cogs & 117 & 61 (52\%) & 35 (30\%) & 27 (23\%) \\
 \cmidrule(l){2-6}
 & \textit{All} & 508 & 329 (65\%) & 229 (45\%) & 145 (29\%) \\
 \bottomrule
 \end{tabular}
\end{table}

\providecommand{\expone}{Exp.~1}\providecommand{\exptwo}{Exp.~2}\providecommand{\expthree}{Exp.~3}
\begin{table*}[t]
 \centering
 \footnotesize
 \setlength{\tabcolsep}{3pt}
 \renewcommand{\arraystretch}{1.15}
 \caption{Robustness of the audit findings, with Cohen's $\kappa$, defined in Eq.~(\ref{eq:kappa}). ET and GL denote the human raters; sol and astra abbreviate the two model names. Unless another comparator is named, message and event percentages report agreement with the pipeline. Chain percentages report the share judged to hold; between-rater chain fractions report agreement on that judgment. Model chain assessments cover every correction and randomly sampled decisions. An evaluation of whether the findings depend on who applies the definitions, not a validation against a ground truth: independent human and model raters are compared with the pipeline and with each other.}
 \label{tab:validation-v6}
 \begin{tabular}{p{3.0cm} p{3.1cm} p{3.4cm} p{3.4cm} p{3.7cm}}
 \toprule
 Step & Human & gpt-5.6-sol & gpt-6-astra & Between raters \\
 \midrule
 {\raggedright Message intent$^{a}$\par} & {\raggedright \textbullet\,ET 53\% ($\kappa$ 0.44), 73\%$^{c}$ ($\kappa$ 0.62), $n{=}60$\newline \textbullet\,GL 57\% ($\kappa$ 0.48), 68\%$^{c}$ ($\kappa$ 0.55), $n{=}60$\par} & {\raggedright \textbullet\,73\% ($\kappa$ 0.68), 80\%$^{c}$ ($\kappa$ 0.71), $n{=}60$\newline \textbullet\,vs ET 52\% ($\kappa$ 0.43), 65\%$^{c}$ ($\kappa$ 0.50), $n{=}60$\newline \textbullet\,vs GL 70\% ($\kappa$ 0.64), 78\%$^{c}$ ($\kappa$ 0.69), $n{=}60$\par} & {\raggedright \textbullet\,77\% ($\kappa$ 0.71), 83\%$^{c}$ ($\kappa$ 0.76), $n{=}60$\newline \textbullet\,vs ET 43\% ($\kappa$ 0.32), 63\%$^{c}$ ($\kappa$ 0.48), $n{=}60$\newline \textbullet\,vs GL 57\% ($\kappa$ 0.47), 72\%$^{c}$ ($\kappa$ 0.59), $n{=}60$\par} & {\raggedright \textbullet\,pipeline raters$^{d}$ $\kappa$ 0.92 / 0.90 / 0.92\newline \textbullet\,ET vs GL 57\% ($\kappa$ 0.47), 72\%$^{c}$ ($\kappa$ 0.60), $n{=}60$\newline \textbullet\,sol vs astra 80\% ($\kappa$ 0.75), 87\%$^{c}$ ($\kappa$ 0.81), $n{=}60$\par} \\
 \midrule
 {\raggedright Coordination type, status and other messages$^{a}$, synthesis vs rest\par} & {\raggedright \textbullet\,ET 60\% ($\kappa$ 0.16, $n{=}68$)\newline \textbullet\,GL 69\% ($\kappa$ 0.36, $n{=}68$)\par} & {\raggedright \textbullet\,79\% ($\kappa$ 0.60, $n{=}68$)\newline \textbullet\,vs ET 60\% ($\kappa$ 0.22, $n{=}68$)\newline \textbullet\,vs GL 63\% ($\kappa$ 0.27, $n{=}68$)\par} & {\raggedright \textbullet\,75\% ($\kappa$ 0.53, $n{=}68$)\newline \textbullet\,vs ET 59\% ($\kappa$ 0.20, $n{=}68$)\newline \textbullet\,vs GL 62\% ($\kappa$ 0.25, $n{=}68$)\par} & {\raggedright \textbullet\,pipeline raters$^{d}$ $\kappa$ 0.86 / 0.86 / 0.91\newline \textbullet\,ET vs GL 62\% ($\kappa$ 0.22, $n{=}68$)\newline \textbullet\,sol vs astra 81\% ($\kappa$ 0.61, $n{=}68$)\par} \\
 \midrule
 {\raggedright Coordination type of ideas, decisions, findings, corrections$^{b}$, synthesis vs rest\par} & {\raggedright \textbullet\,ET 71\% ($\kappa$ 0.23, $n{=}51$)\newline \textbullet\,GL 59\% ($\kappa$ 0.19, $n{=}51$)\par} & {\raggedright \textbullet\,96\% ($\kappa$ 0.73, $n{=}51$)\newline \textbullet\,vs ET 71\% ($\kappa$ 0.21, $n{=}51$)\newline \textbullet\,vs GL 55\% ($\kappa$ 0.11, $n{=}51$)\par} & {\raggedright \textbullet\,98\% ($\kappa$ 0.88, $n{=}51$)\newline \textbullet\,vs ET 73\% ($\kappa$ 0.27, $n{=}51$)\newline \textbullet\,vs GL 57\% ($\kappa$ 0.15, $n{=}51$)\par} & {\raggedright \textbullet\,ET vs GL 69\% ($\kappa$ 0.38, $n{=}51$)\newline \textbullet\,sol vs astra 98\% ($\kappa$ 0.85, $n{=}51$)\par} \\
 \midrule
 {\raggedright Event type (idea, decision, finding, correction)$^{b}$\par} & {\raggedright \textbullet\,66\% ($\kappa$ 0.46, $n{=}86$)\newline \textbullet\,operational vs rest 83\% ($\kappa$ 0.65, $n{=}86$)\par} & {\raggedright \textbullet\,91\% ($\kappa$ 0.84, $n{=}86$)\newline \textbullet\,vs GL 69\% ($\kappa$ 0.51, $n{=}86$)\newline \textbullet\,operational vs rest 95\% ($\kappa$ 0.91, $n{=}86$)\par} & {\raggedright \textbullet\,91\% ($\kappa$ 0.84, $n{=}86$)\newline \textbullet\,vs GL 72\% ($\kappa$ 0.56, $n{=}86$)\newline \textbullet\,operational vs rest 95\% ($\kappa$ 0.91, $n{=}86$)\par} & {\raggedright \textbullet\,pipeline raters$^{d}$ $\kappa$ 0.97 / 0.95 / 0.96\newline \textbullet\,sol vs astra 95\% ($\kappa$ 0.92, $n{=}86$)\par} \\
 \midrule\midrule
 {\raggedright Key change, \expone{}: single $\to$ multi-objective$^{e}$\par} & {\raggedright 12/18 (67\%)\par} & {\raggedright 17/18 (94\%)\par} & {\raggedright 15/18 (83\%)\par} & {\raggedright \textbullet\,human vs sol 13/18\newline \textbullet\,human vs astra 9/18\newline \textbullet\,sol vs astra 14/18\par} \\
 \midrule
 {\raggedright Key change, \exptwo{}: UCB $\to$ radical gradient$^{e}$\par} & {\raggedright 5/6 (83\%)\par} & {\raggedright 5/6 (83\%)\par} & {\raggedright 5/6 (83\%)\par} & {\raggedright \textbullet\,human vs sol 4/6\newline \textbullet\,human vs astra 4/6\newline \textbullet\,sol vs astra 6/6\par} \\
 \midrule
 {\raggedright Correction chains$^{b,e}$\par} & {\raggedright 18/21 (86\%)\par} & {\raggedright 330/392 (84\%)\par} & {\raggedright 310/392 (79\%)\par} & {\raggedright \textbullet\,on the human's chains: sol 19/21, astra 17/21\newline \textbullet\,human vs sol 16/21\newline \textbullet\,human vs astra 16/21\newline \textbullet\,sol vs astra 350/392\par} \\
 \midrule
 {\raggedright Decision chains, random$^{b,e}$\par} & {\raggedright 8/13 (62\%)\par} & {\raggedright 390/450 (87\%)\par} & {\raggedright 377/450 (84\%)\par} & {\raggedright \textbullet\,on the human's chains: sol 12/13, astra 12/13\newline \textbullet\,human vs sol 7/13\newline \textbullet\,human vs astra 7/13\newline \textbullet\,sol vs astra 411/450\par} \\
 \midrule
 {\raggedright Same 106 random chains labeled twice$^{b,e}$\par} & {\raggedright ---\par} & {\raggedright same verdict 99/106 (93\%)\par} & {\raggedright same verdict 104/106 (98\%)\par} & {\raggedright ---\par} \\
 \bottomrule
 \end{tabular}
 \par\smallskip
 {\raggedright $^{a}$\expthree{} messages. $^{b}$All three experiments. $^{c}$The seven intents in four coarser groups: idea
 and decision; finding, status and correction; question; other. $^{d}$Cohen's $\kappa$ of the pipeline's two raters before adjudication by the third, over all messages for message classifications and all labeled events for event focus, using the full label set; values are ordered \expone{} / \exptwo{} / \expthree. $^{e}$Strict: a chain holds
 only if every link holds, including who is credited with each step.\par}
\end{table*}

\end{document}